\documentclass[lettersize,journal]{IEEEtran}
\usepackage{amsmath,amsfonts}
\usepackage{algorithmic}
\usepackage{algorithm}
\usepackage{array}
\usepackage[caption=false,font=normalsize,labelfont=sf,textfont=sf]{subfig}
\usepackage{textcomp}
\usepackage{stfloats}
\usepackage{url}
\usepackage{verbatim}
\usepackage{graphicx}
\usepackage{pifont}
\usepackage{multirow}
\usepackage{booktabs}
\usepackage{xcolor}
\usepackage{amsmath}
\usepackage{makecell}
\usepackage{tcolorbox}
\usepackage{hyperref}
\usepackage{colortbl}
\usepackage{cite}
\newcommand{\ie}{\textit{i.e.}}
\newcommand{\eg}{\textit{e.g.}}
\begin{document}

\title{HVI-CIDNet+: Beyond Extreme Darkness \\ for Low-Light Image Enhancement}

\author{Qingsen Yan, Kangbiao Shi, Yixu Feng, \\Tao Hu, Peng Wu, Guansong Pang, and Yanning Zhang$^{\ast}$\thanks{$^{\ast}$Corresponding authors: Yanning Zhang.},~\IEEEmembership{Fellow,~IEEE}
\thanks{This work is supported by NSFC of China under Grant 62301432 and Grant 6230624, the Natural
Science Basic Research Program of Shaanxi under Grant 2023-JC-QN-0685
and Grant QCYRCXM-2023-057, the Fundamental Research
Funds for Central Universities, and Guangdong Basic and Applied Basic Research Foundation 2025A1515011119.}
\thanks{Qingsen Yan is with the School of Computer Science and Shenzhen Research Institute, Northwestern Polytechnical University, Xi’an, China. (e-mail: qingsenyan@nwpu.edu.cn).}
\thanks{Kangbiao Shi, Yixu Feng, Tao Hu, Peng Wu, and Yanning Zhang are with the School of Computer Science, Northwestern Polytechnical University, Xi’an, China. (e-mail: 18334840904@163.com; yixu-nwpu@mail.nwpu.edu.cn; taohu@mail.nwpu.edu.cn; pengwu@nwpu.edu.cn; ynzhang@nwpu.edu.cn).}
\thanks{Guansong Pang is with the School of Computing and Information Systems, Singapore Management University. (e-mail: gspang@smu.edu.sg).}
}


\maketitle

\begin{abstract}
Low-Light Image Enhancement (LLIE) aims to restore vivid content and details from corrupted low-light images.
However, existing standard RGB (sRGB) color space-based LLIE methods often produce color bias and brightness artifacts due to the inherent high color sensitivity.
While Hue, Saturation, and Value (HSV) color space can decouple brightness and color, it introduces significant red and black noise artifacts.
To address this problem, we propose a new color space for LLIE, namely Horizontal/Vertical-Intensity (HVI), defined by the HV color map and learnable intensity. The HV color map enforces small distances for the red coordinates to remove red noise artifacts, while the learnable intensity compresses the low-light regions to remove black noise artifacts. 
Additionally, we introduce the Color and Intensity Decoupling Network+ (HVI-CIDNet+), built upon the HVI color space, to restore damaged content and mitigate color distortion in extremely dark regions. 
Specifically, HVI-CIDNet+ leverages abundant contextual and degraded knowledge extracted from low-light images using pre-trained vision-language models, integrated via a novel Prior-guided Attention Block (PAB). 
Within the PAB, latent semantic priors can promote content restoration, while degraded representations guide precise color correction, both particularly in extremely dark regions through the meticulously designed cross-attention fusion mechanism.
Furthermore, we construct a Region Refinement Block that employs convolution for information-rich regions and self-attention for information-scarce regions, ensuring accurate brightness adjustments.
Comprehensive results from benchmark experiments demonstrate that the proposed HVI-CIDNet+ outperforms the state-of-the-art methods on 10 datasets. The code is available at \href{https://github.com/shikangbiao/CIDNet_extension}{\textcolor{red}{https://github.com/shikangbiao/CIDNet\_extension}}.
\end{abstract}

\maketitle
\section{Introduction}
\label{sec:intro}
\begin{figure*}[htp]
    \centering
    \includegraphics[width=1\linewidth]{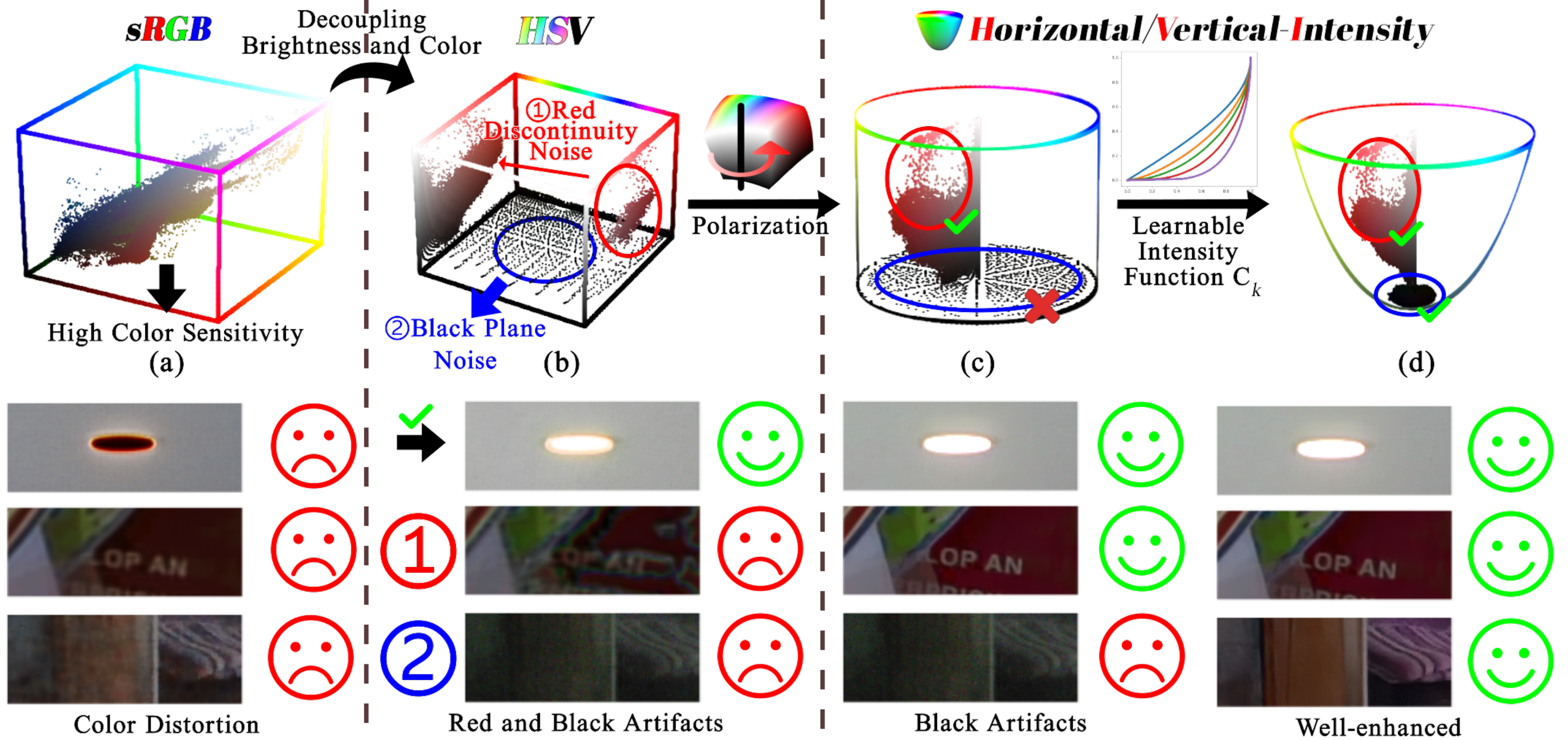}
    \caption{The top row illustrates the process of transforming images from the sRGB color space, via HSV, to the HVI color space. The bottom row presents the corresponding test results. The sRGB color space is known for its high color sensitivity, often causing color distortions in test images. By decoupling brightness and color to obtain the HSV color space, the illumination enhancement appears normalized. However, this transformation introduces varying levels of red discontinuity and black regions, which subsequently cause artifacts in the enhanced images. Introducing polarization to HSV color space ensures continuity in the red regions. Introducing a learnable intensity function $\mathbf{C}_k$, on the other hand, helps collapse the black regions, yielding the HVI color space with optimal image enhancement.}
    \label{fig:1}
\end{figure*} 
\IEEEPARstart{I}{maging} sensors often capture weak light signals with severe noise under low-light conditions, resulting in poor visual quality for low-light images.
Low-Light Image Enhancement (LLIE) can obtain high-quality images from such degraded images, 
which improves the image brightness and suppresses noise ~\cite{2022LLE}, meanwhile, restoring damaged content and mitigating color distortion in extremely dark regions~\cite{SNR-Aware,xu2023low}.

Most existing LLIE approaches are based on the standard RGB (sRGB) color space \cite{KinD, EnGAN,Zero-DCE,Han_ECCV24_GLARE,10543170,zhang2021beyond,li2021learning,wu2025interpretable}.
However, the image brightness exhibits a strong coupling with the color of the three sRGB channels, \textit{a.k.a.} high color sensitivity \cite{yan2025hvi,gevers2012color,lee2024rethinking}, 
causing an obvious color distortion of the restored image in these LLIE methods \cite{RetinexFormer,EnGAN}, as shown in Fig.~\ref{fig:1} (a).
Inspired by Kubelka-Munk theory \cite{gevers2012color}, recent methods \cite{li2020low,zhou2023low,zhang2021better} have sought to transform images from the sRGB color space to the Hue, Saturation and Value (HSV) color space. 
These methods can improve brightness more accurately, but they amplify local color space noise \cite{gevers2012color}, introducing severe artifacts in the results. 
As illustrated in Fig. \ref{fig:1} (b), the transformation from sRGB to the HSV disrupts the continuity of the red (\textcolor{red}{\ding{172} Red Discontinuity Noise}) and black (\textcolor{blue}{\ding{173} Black Plane Noise}) color, resulting in increased Euclidean distances for similar color and the obvious artifacts in the final images (see the zoomed in images of \textcolor{red}{\ding{172}} and \textcolor{blue}{\ding{173}}). These two types of noise will cause severe artifacts in the enhancement of red- or black-dominated images.
Moreover, regardless of the color space employed, effective pixel values fall to near zero and mix with sensor noise in extremely dark regions, which completely results in the loss of semantic information and hinders complete content reconstruction~\cite{SNR-Aware,xu2023low,wang2024zero,ma2024region}. 
Meanwhile, colors in these regions become nearly indistinguishable due to minimal pixel value variation, making precise color correction impossible~\cite{wu2023learning}.

To address red and black noise artifacts, we introduce a new color space named Horizontal/Vertical-Intensity (HVI) 
to improve the image brightness while reducing the impact of noise.
The key intuition is that minimizing color space noise by reducing Euclidean distances in similar colors. To this end, we polarize in the Hue and Saturation plane to enforce smaller distances for similar red point coordinates, which eliminates the red discontinuity noise in the HSV color space (see Fig. \ref{fig:1} (c)). 
By polarizing the Hue and Saturation plane, this transformation produces HV color map that retains rich and detailed color information.
For the black plane noise issue, we introduce a trainable darkness density parameter $k$ and its corresponding adaptive intensity collapse function $C_k$, which compresses the radius of low-light regions to zero, with the flexibility to gradually expand to the value of one as the intensity increases, as illustrated in Fig. \ref{fig:1} (d). This operation effectively helps remove black noise artifacts while maintaining the primary image appearance.
Building on the novel HVI color space, we further propose 
Color and Intensity Decoupling Network+ (HVI-CIDNet+)
to address content deficiency and color distortion in extremely dark regions. 
Specifically, HVI-CIDNet+ consists of the HV-branch and the intensity-branch (I-branch) to take full use of the color and intensity information.
To achieve precise content restoration and color correction, HVI-CIDNet+ employs image encoder and distortion encoder to introduce abundant contextual and degraded knowledge from pre-trained visual-language models (VLMs)~\cite{luo2023controlling}, which are integrated through the Prior-guided Attention Block (PAB).
Within the PAB, the I‑branch leverages latent semantic priors to extract accurate brightness features, while the HV‑branch utilizes degraded representations to learn color features without distortion. This complementary design enables the cross-attention mechanism to effectively integrate brightness and color features, facilitating bidirectional information interplay between dual-branch.
Furthermore, we design region refinement block, which uses a context predictor and a mask predictor to generate spatial masks that partition feature maps into information-scarce regions and information-rich regions, processed by convolution and attention branches, respectively. 
Comprehensive results from quantitative and qualitative experiments show that our approach outperforms various types of state-of-the-art (SOTA) methods on different metrics across 10 datasets. 

The main contributions of our work are as follows.
\begin{itemize}
    \item We introduce a new HVI color space for the LLIE task, which is uniquely defined by the HV color map and trainable intensity. This offers an effective tool that eliminates the color space noise arising from the HSV color space, effectively enhancing the brightness of low-light images.
    \item We propose HVI-CIDNet+ to restore damaged content and mitigate color distortion in extremely dark regions, utilizing latent semantic priors and degraded representations.
    \item We design the Prior-guided Attention Block (PAB) to enable mutual feature guidance between dual-branch, ensuring accurate recovery of both color and illumination.
    \item We construct the region refinement block to partition low-light images into distinct regions, respectively improving brightness in information-rich regions and information-scarce regions.
\end{itemize}

This work is an extension of our conference version~\cite{yan2025hvi} that has been published in Computer Vision and Pattern Recognition (CVPR 2025). Compared with the CVPR version, we have introduced a significant amount of new materials.
1) Considering that extremely dark regions often suffer from severe content deficiency and color distortion, we propose an advanced version of CIDNet, called HVI-CIDNet+. The HVI-CIDNet+ introduces abundant contextual and degraded knowledge from pre-trained VLMs, effectively guiding content restoration and ensuring accurate color correction. 
2) We design Prior-guided Attention Block to achieve the feature interaction under the guidance of degraded representations and latent semantic priors, which enhances the interplay between color and brightness information.
3) To address the severe degradation of content and details caused by uneven brightness in low-light images, we construct the Region Refinement Block, which respectively restores content and enhances details in information-scarce regions and information-rich regions, effectively optimizing brightness adjustment.
4) Correspondingly, we perform more experiments on HVI-CIDNet+ to demonstrate its advantages. Specifically, the enriched qualitative and quantitative experiments on 10 datasets are carried out, and more ablation studies are also added to verify the superiority of HVI-CIDNet+.
\section{Related Work}
\subsection{Color Model}
\textit{RGB.}
Due to the same principle as visual recognition by the human eye, the sRGB color space is widely used in LLIE. However, the image brightness and color exhibit a strong interdependence with the three channels in sRGB \cite{gevers2012color}. A slight disturbance in the color space will cause an obvious variation in both the brightness and color of the restored image. Thus, sRGB is not a desired color space for the LLIE task.

\textit{HSV and YCbCr.} 
The HSV color space represents points in an RGB color model with a cylindrical coordinate system \cite{Foley1982FundamentalsOI}. Indeed, it does decouple the brightness and color of the image from the sRGB channels. However, the red color discontinuity and black plane noise pose significant challenges when we enhance the images in HSV color space, resulting in the emergence of various obvious artifacts. To circumvent issues related to HSV color space, some methods \cite{Bread,brateanu2024lytnet} also transform sRGB images into the YCbCr color space, which has an illumination axis (Y) and reflect-color-plane (CbCr). Although it solved the hue dimension discontinuity problem of HSV color space, the Y axis is still partially coupled with the CbCr plane, leading to severe color shifts.

\textit{Retinex Theory.} 
The Retinex-based color model decomposes an image into reflectance and illumination components,  assuming that reflectance remains invariant across varying lighting conditions. 
This color model transforms brightness enhancement into illumination estimation and noise suppression problem. 
Retinex-based methods~\cite{RetinexNet,URetinexNet,RetinexFormer,luo2024unsupervised,wang2013naturalness,fu2015probabilistic} can effectively enhance contrast and improve brightness under moderately low-light conditions. 
However, accurately estimating reflectance and illumination components with neural networks is challenging, as reflectance map inherently includes brightness information, and increasing brightness can distort color information. This leads to noise artifacts in extremely dark regions, limiting the direct application in the LLIE tasks.

\textit{Kubelka-Munk Theory.} 
The color model~\cite{gevers2012color} describes light propagation in the medium by accounting for scattering and absorption processes. QuadPrior~\cite{wang2024zero} leverages this model to characterize illumination factors in low-light images, proposing the physical quadruple prior to extract illumination-invariant features. These features then guide the network to restore low‑light images, showing obvious robustness across various scenarios. 
However, the Kubelka–Munk formulation is fundamentally a grayscale model, which falls short in describing colors, leading to color biases in the restored images.
\subsection{Low-Light Image Enhancement}
\textit{Single-stage Methods.}
Single-stage deep learning approaches~\cite{RetinexNet,LLNet,KinD,EnGAN,SNR-Aware,10244055,9809998,LLFlow,wang2023flow,wang2023residual,wang2023fourllie} has been widely used in the LLIE task. Existing methods propose distinct solutions to address the aforementioned issues. 
For instance, RetinexNet \cite{RetinexNet} enhances images by decoupling illumination and reflectance based on Retinex theory. 
Bread \cite{Bread} decouples the entanglement of noise and color distortion by using YCbCr color space. Furthermore, they designed a color adaption network to tackle the color distortion issue left in the restored images. 
SNRNet \cite{SNR-Aware} integrates signal-to-noise-ratio-aware transformers to perform long-range operations on extremely low signal-to-noise ratio image regions and short-range operations on other regions.
Still, the RetinexNet, SNRNet, and Bread can show inaccurate control in terms of brightness and color in extremely dark regions. 

\textit{Methods Incorporating Pre-Trained Models.}
These methods utilize knowledge from large-scale datasets, such as vision-language or segmentation models.
CLIP-LIT~\cite{liang2023iterative} leverages the potential of the CLIP~\cite{radford2021learning} through iterative prompt learning to optimize the enhancement network, significantly improving the visual quality of degraded images.
Similarly, SKF~\cite{wu2023learning} employs a pre-trained HRNet~\cite{wang2020deep} as a semantic knowledge base to provide prior information, effectively enhancing content and preserving details in low-light images.
However, these approaches often struggle with precise color correction in extremely dark conditions, as standard color spaces like sRGB or HSV introduce artifacts during processing. Additionally, recovering details in information-scarce regions remains challenging due to complete loss of information.

\textit{Diffusion-based Methods.}
With the advancement of Denoising Diffusion Probabilistic Models (DDPMs)~\cite{Ho2020Denosing}, diffusion-based generative approaches~\cite{jiang2024lightendiffusion,yi2023diff,zhou2023pyramid,10740586,shang2024multi,jiang2023low} can generate more accurate and appropriate images by gradually removing noise, showing realistic details in the restored image. 
PyDiff~\cite{zhou2023pyramid} employs a pyramid diffusion approach and a global corrector to address the slow processing and global degradation issues of traditional diffusion models in the LLIE task, significantly improving enhancement quality and efficiency.
LightenDiffusion \cite{jiang2024lightendiffusion} embeds the Retinex decomposition mechanism in the latent space for reconstruction of relevant reflectance maps. 
However, these diffusion models often produce unrealistic content in extremely dark regions, thereby resulting in poor visual quality.
\subsection{Vision-Language Models}
Recent research \cite{jia2021scaling,li2022blip,ai2024lora} results have shown that the application of pre-trained vision-language models (VLMs) with generic visual and textual representations has great potential to improve downstream tasks.
Among them, CLIP~\cite{radford2021learning}, as a classical VLM, usually consists of the text encoder and the image encoder, utilize large-scale image-text comparison learning to achieve cross-modal feature alignment, demonstrating impressive zero-shot and few-shot capabilities across various high-level semantic tasks~\cite{wanghard,zhou2023zegclip,wang2023improving}. 

However, in low-level vision tasks, CLIP's potential have been relatively less explored. 
DA-CLIP~\cite{DACLIP} is the first to incorporate CLIP into all-in-one image restoration, employing contrastive learning with image-text pairs to fine-tune CLIP.
In this work, we leverage DA-CLIP's visual representation capabilities to effectively capture latent semantic priors and degraded representations from low-light images, improving the quality of the restored image.
\section{HVI Color Space}
The HVI color space is built upon the HSV color space, which is proposed to address the color space noise issues arising from the HSV color space. The key intuition in HVI color space is that the restored images should have good perceptual quality for respective colors, \textit{\ie} similar colors have small Euclidean distances.
Below we introduce the HVI transformation in detail, where the HSV color space is first applied to decouple the brightness and color information of input images, which could cause color space noise (\textit{\eg} red discontinuity noises and black plane noises). We then introduce our proposed polarized HS operations and learnable intensity collapse function in HVI color space to effectively address these issues.
\subsection{Color Space Noises in HSV}
\textit{Intensity Map.}
In the task of LLIE, one crucial aspect is accurately estimating the illumination intensity map of the scene from a sRGB input image. Previous methods \cite{RetinexNet, PairLIE, KinD} largely rely on the Retinex theory \cite{land1977retinex}, using deep learning to directly generate the corresponding normal-light map. While this approach aligns with statistical principles, it often struggles to fit physical laws and human perception \cite{gevers2012color}, resulting in limited generalizability \cite{wang2024zero}. Therefore, we instead refer to the Max-RGB theory \cite{land1977retinex} to estimate the intensity map, \textit{a.k.a.}, Value in HSV, rather than using neural networks to generate it. According to Max-RGB theory, for each individual pixel $x$, we can estimate the intensity map of an image $\mathbf{I}_{max} \in \mathbb{R}^{\mathrm{H \times W}} $ as follows:
\begin{equation}
\mathbf{I}_{max}(x) =\max_{\mathbf{c}\in \{R,G,B\}} (\mathbf{I_{c}}(x)).
\label{eq:1}
\end{equation}
The intensity map then goes through the sRGB-HSV transformation that can lead to various red and black noise artifacts. We introduce these separately as follows.

\textit{Hue/Saturation Plane.}
Real-world low-light images often contain significant noise, making its identification and removal a key challenge in the LLIE task. Recent studies \cite{RetinexNet,yi2023diff} indicate that the noise in low-light images is a primary cause to shifts in Hue and Saturation, \textit{a.k.a.} a general case of Reintex theory \cite{wang2024zero}, while having minimal impact on light intensity. 
Therefore, decoupling the sRGB color space, known for its high color sensitivity, can be advantageous for the LLIE task. By leveraging pixel-based photometric invariance \cite{gevers2012color} and dichromatic reflection modeling \cite{shafer1985using}, sRGB can be decoupled into illuminance and chromatic components, yielding the HSV (Hue/Saturation-Value) color space.
In this representation, the Value map ($\mathbf{V}$) component corresponds to light intensity map ($\mathbf{V}=\mathbf{I}_{max}$), while the HS plane forms a chromaticity plane independent of illuminance constraints. Specifically, the transformation of sRGB image to Saturation map ($\mathbf{S}$) is defined as follows:
\begin{equation}
\mathbf{s}  = \begin{cases}
0, &  \mathbf{I}_{max}   = 0 \\
\frac{\Delta}{\mathbf{I}_{max} }, &  \mathbf{I}_{max}  \ne 0 \\
\end{cases} \\
,
\end{equation}
where $\Delta=\mathbf{I}_{max}-min(\mathbf{I}_c)$ and $\mathbf{s}$ is any pixel in $\mathbf{S}$. The Hue map ($\mathbf{H}$) is formulated as:
\begin{equation}
\mathbf{h}= \begin{cases}
0, &\text{if } \mathbf{s} = 0 \\
\frac{\mathbf{I_{G}} - \mathbf{I_{B}}}{\Delta} \mod{6},  &\text{if } \mathbf{I}_{max}  = \mathbf{I_{R}} \\
2+\frac{\mathbf{I_{B}} - \mathbf{I_{R}}}{\Delta},  &\text{if } \mathbf{I}_{max}   = \mathbf{I_{G}} \\
4+\frac{\mathbf{I_{R}} - \mathbf{I_{G}}}{\Delta},  &\text{if } \mathbf{I}_{max}   = \mathbf{I_{B}}
\end{cases}
,
\label{eq:hue}
\end{equation}
where $\mathbf{h}$ is any pixel in $\mathbf{H}$. 

\textit{Color Space Noises.} 
Converting an sRGB image to HSV color space effectively decouples brightness from color, enabling more accurate color denoising and more natural illuminance recovery. However, this transformation also amplifies noise in the red and dark regions \cite{gevers2012color}, which are critical for the LLIE task. As illustrated in Fig. \ref{fig:1} (b), enhancing the image brightness within the HSV color space yields a more balanced brightness level. However, excessive noise in the red discontinuity and the black plane introduces significant artifacts, particularly in red and dark regions of output image, which greatly degrade the perceptual quality. 

To address this issue, we propose the HVI color space as follows, which effectively preserves the decoupling of brightness and color while minimizing noise artifacts.
\begin{figure*}[!htp]
    \centering
    \includegraphics[width=1\linewidth]{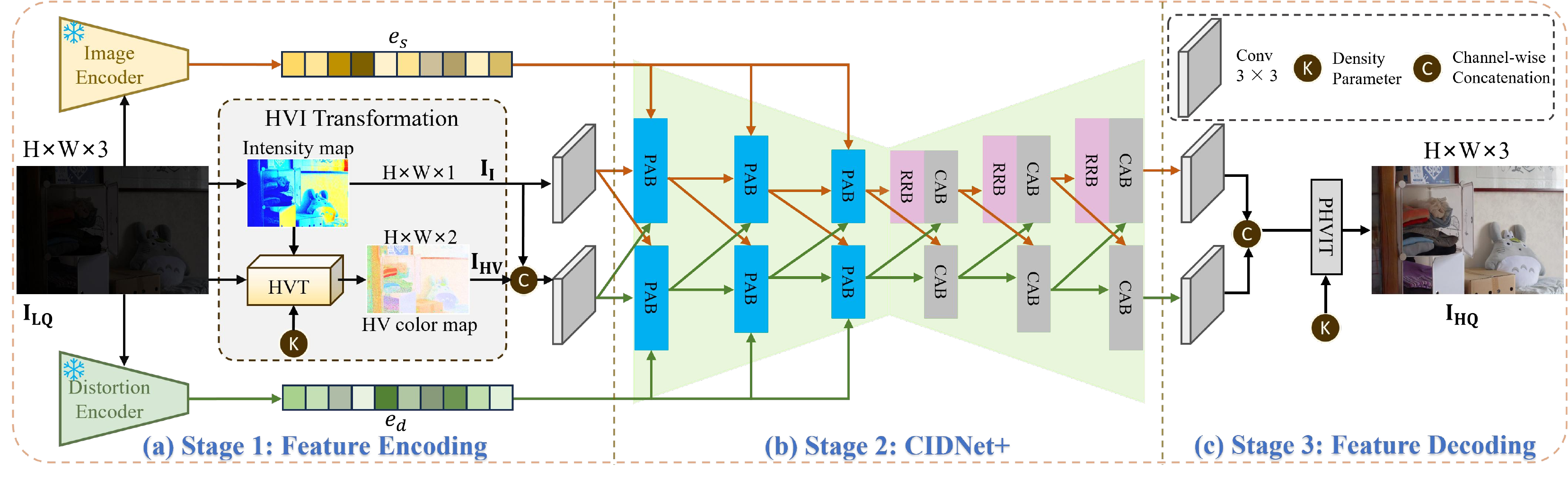}
    \caption{Overall of the proposed HVI-CIDNet+. In Stage 1 (left), the input low-light sRGB image is converted to the HVI color space, yielding HV color map and intensity map. Meanwhile, we extract latent semantic priors and degraded representations from pre-trained VLMs. 
    For Stage 2 (center), the CIDNet+ comprises two nested U‑shaped subnetworks: an HV‑branch for color processing and an I‑branch for brightness enhancement,  integrated via the novel Prior-guided Attention Block (PAB) and the Region Refinement Block (RRB). 
    In Stage 3 (right), the light-up HVI map are mapped back to sRGB via the Perceptual-inverse HVI Transformation (PHVIT) to produce the sRGB-enhanced image.}
    \label{fig:CIDNet-pipeline}
\end{figure*}
\subsection{Horizontal/Vertical Plane with Polarized HS and Collapsible Intensity}
\label{sec:HVI} 
To address the color space noise issue, our primary approach is to ensure that more similar colors exhibit smaller Euclidean distances. Along the Hue axis, the red color appears identically at both $\mathbf{h}=0$ and $\mathbf{h}=6$, due to the modular arithmetic of Hue-axis, which splits the same color across two ends of the spectrum. In particular, to address the red discontinuity issue, we apply polarization to the each pixel in $\mathbf{H}$, obtaining orthogonal $h= \cos (\frac{\pi \mathbf{h}}{3})$ and $v= \sin (\frac{\pi \mathbf{h}}{3})$.
When the Hue axis is polarized, it forms an angle within the orthogonalized $h-v$ plane, with $\mathbf{s}$ representing the distance from the origin.

For the black plane noise issue, we aim to collapse regions of low-light intensity while preserving those with higher intensity. However, the optimal extent of collapse varies across different datasets and networks. Therefore, it is important to make this region adaptively collapsible through a learning process. To achieve this, we introduce an adaptive intensity collapse function $\mathbf{C}_k$ as follows:
\begin{equation}
\mathbf{C}_k(x)=\sqrt[k]{\sin (\frac{ \pi \mathbf{I}_{max}(x) }{2} )+\mathcal{\varepsilon} },
\label{eq:2}
\end{equation}
where $k\in \mathbb{Q^+}$ is a trainable parameter to control the dark color point density, and a small  $\mathcal{\varepsilon}=1\times10^{-8}$ is used to avoid gradient explosion. Essentially, $\mathbf{C}_k$ serves a radius mapping function, with smaller $\mathbf{C}_k$ corresponding to smaller radius or lower intensity values. Thus, black points are clustered together as $\mathbf{C}_k$ decreases. We then formalize the Horizontal ($\mathbf{\hat{H}}$) map and Vertical ($\mathbf{\hat{V}}$) map as:
\begin{equation}
\begin{split}
    \mathbf{\hat{H}} &= \mathbf{C}_k \odot  \mathbf{S}  \odot H,\\
    \mathbf{\hat{V}} &= \mathbf{C}_k \odot  \mathbf{S}  \odot V,
\end{split}
\label{eq:7}
\end{equation}
where $h \in H$, $v \in V$, and $\odot$ denotes the element-wise multiplication. $\mathbf{\hat{H}}$, $\mathbf{\hat{V}}$, and $\mathbf{I}_{max}$ can be concatenated to form an HVI map.

Thanks to these operations, the HVI builds a strong color space that maintains the advantages of HSV color space while effectively improving the image brightness and suppresses noise.
\section{HVI-CIDNet+}
To restore severely degraded information in extremely dark regions, we further present HVI-CIDNet+, which effectively alleviates content deficiency and color distortion in these regions, accurately learning the mapping relationship between low-light images and normal-light images in the HVI color space.
\subsection{Overview}
Fig. \ref{fig:CIDNet-pipeline} illustrates the overall framework of the proposed HVI-CIDNet+, which is divided into three consecutive stages.
Stage 1 is the feature encoding, which comprises HVI transformation and extraction of abundant contextual and degraded knowledge from pre-trained Vision-Language Models (VLMs).
Stage 2 is the CIDNet+, which employs a dual‐branch enhancement network to restore low-light images.
Stage 3 is the feature decoding, which is essential for transforming light-up HVI map back into the sRGB color space to generate sRGB-enhanced image.

We first introduce the feature encoding in Sec.~\ref{sec:feature_encoding}, which transforms low-light images from the sRGB color space to the HVI color space through HVI transformation, yielding the HV color map and the intensity map. Meanwhile, degraded representations and latent semantic priors extracted from pre-trained DA-CLIP guide the training of CIDNet+. 
Then we detail the CIDNet+ in Sec.~\ref{sec:CIDNet+}, which consists of the HV-branch and the intensity-branch (I-branch). In this stage, the HV-branch concentrates on suppressing noise and mitigating color distortion in extremely dark regions, while I-branch respectively restores content and enhances details in information-scarce regions and information-rich regions, effectively optimizing brightness adjustment.
Finally, we describe the feature decoding process in Sec.~\ref{sec:feature_decoding}.
\subsection{Feature Encoding}
\label{sec:feature_encoding}
As described in Sec. \ref{sec:HVI}, we first apply the HVI transformation to decompose the sRGB low-light image $\mathbf{I_{LQ}}$ into an HV color map $\mathbf{I_{HV}}$ and an intensity map $\mathbf{I_I}$.
Specifically, we first calculate the $\mathbf{I_I}$ using Eq. \ref{eq:1}, which is $\mathbf{I_{I}}=\mathbf{I}_{max}$. Subsequently, we utilize the $\mathbf{I_I}$ and the $\mathbf{I_{LQ}}$ to generate the $\mathbf{I_{HV}}$ using Eq. \ref{eq:7}. Furthermore, a trainable density-$k$ is employed to adjust the color point density of the low-intensity color plane.

As shown in Fig.~\ref{fig:CIDNet-pipeline} (a),  we utilize image encoder to extract latent semantic priors $\mathbf{e}_s$, which guide I-branch to extract accurate brightness features, effectively restoring content and improving illumination. 
Meanwhile, we use the distortion encoder to introduce the degraded representations $\mathbf{e}_d$, which promotes denoising and precise color correction in the HV‑branch.

Therefore, this stage generates the four essential inputs for CIDNet+: $\mathbf{I_{HV}}$, $\mathbf{I_I}$, $\mathbf{e}_s$, and $\mathbf{e}_d$, which collectively enable effective enhancement in the subsequent stage.
\begin{figure}
    \centering
    \includegraphics[width=0.85\linewidth]{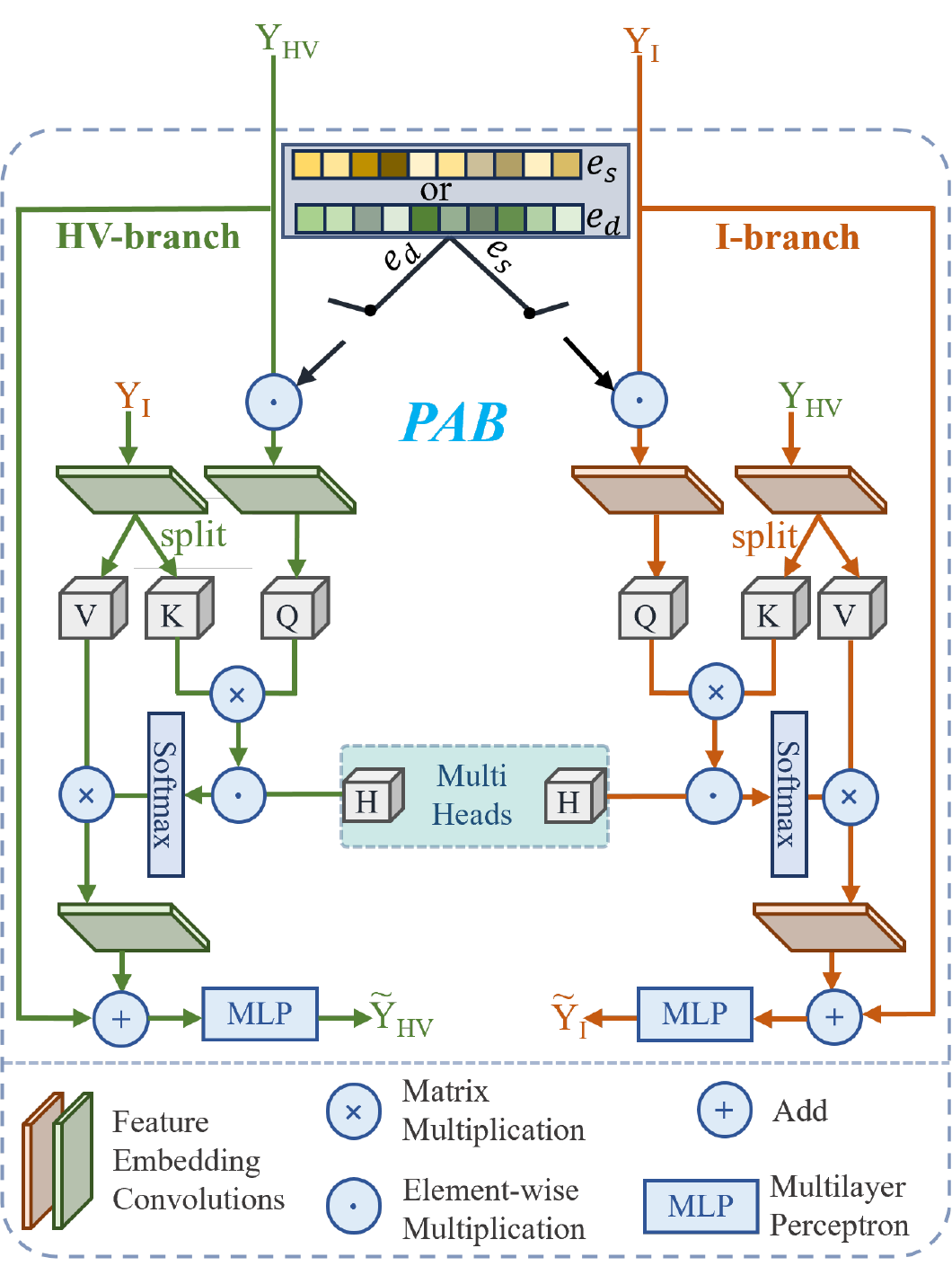}
    \caption{The architecture of Prior-guided Attention Block (PAB). The PAB uses cross-attention to integrate latent semantic priors $\mathbf{e}_s$ and degraded representations $\mathbf{e}_d$ into a dual-branch network for content restoration and color correction in extremely dark regions.}
    \label{fig:LCA}
\end{figure}
\subsection{CIDNet+}
\label{sec:CIDNet+}
As described in Sec. \ref{sec:intro}, the LLIE task can be decomposed into two sub-tasks: (1) noise removal and brightness enhancement and (2) content restoration and color correction in extremely dark regions.
Since these two sub-tasks follow distinct statistical patterns \cite{2022LLE}, inspired by Retinex-based methods \cite{RetinexNet,URetinexNet,PairLIE},
we divide CIDNet+ into two parallel branches, which consists of two parallel U‑shaped networks. As shown in Fig. \ref{fig:CIDNet-pipeline} (b), each UNet incorporates three Prior-guided Attention Blocks (PABs) and three Cross-Attention Block (CABs)~\cite{yan2025hvi}.
Furthermore, the I‑branch features three Region Refinement Blocks (RRBs) inserted before each CAB to respectively restore content and enhance details in information-scarce regions and information-rich regions, accurately adjusting brightness of different regions.

\textit{PAB.} To facilitate feature interaction between the dual-branch, we design the Prior-guided Attention Block (PAB), allowing each branch to selectively leverage the other’s strengths under abundant contextual and degraded knowledge guidance.

As shown in Fig. \ref{fig:LCA}, the latent semantic priors $\mathbf{e}_s$ contains rich semantic information of low-light image $\mathbf{I_{LQ}}$, which is injected into the I-branch. 
By incorporating the $\mathbf{e}_s$ into the query($\mathbf{Q}$) of the cross-attention mechanism in the I-branch, the network can better understand the semantic context of the $\mathbf{I_{LQ}}$, enabling more accurate restoration of brightness features and enhancement of content in extremely dark regions. 
Meanwhile, the degraded representations $\mathbf{e}_d$ captures the degradation characteristics of $\mathbf{I_{LQ}}$ and is injected into the HV-branch to assist in denoising and color correction. When combined with the $\mathbf{Q}$ of the cross-attention mechanism in the HV-branch, the $\mathbf{e}_d$ guides the network to distinguish between noise and useful color information, enabling to more effectively suppress noise and mitigate color distortion in extremely dark regions.

Next, we illustrate the PAB by using its implementation in the I‑branch as an example.
Given the input feature of the I-branch, $\mathbf{Y_I} \in \mathbb{R}^{\hat{H} \times \hat{W} \times \hat{C}}$, the input feature of the HV-branch, $\mathbf{Y_{HV}} \in \mathbb{R}^{\hat{H} \times \hat{W} \times \hat{C}}$, along with the latent semantic priors $\mathbf{e}_s \in \mathbb{R}^{512}$ and the degraded representations $\mathbf{e}_d \in \mathbb{R}^{512}$,
PAB first project the $\mathbf{e}_s$ into a spatial feature map ($\mathbf{M}$). Then queries elements ($\mathbf{Q}$) generates by both $\mathbf{M}$ and $\mathbf{Y_{I}}$. Formally:
\begin{equation}
\begin{gathered}
\mathbf{M} = \text{Linear}_{\text{proj}}(\mathbf{e_s}) \in \mathbb{R}^{\hat{C} \times \hat{H} \times \hat{W}}, \\
\mathbf{Q} = W_s^Q W_d^Q(\mathbf{M} \odot \mathbf{Y_{I}}),
\end{gathered}
\end{equation}
Where $\odot$ denotes the element-wise multiplication. $W_s^{(\cdot)}$ is a group conv $3 \times 3$ layer and $W_d^{(\cdot)}$ is a depth-wise conv $1 \times 1$ layer. Meanwhile, PAB synthesizes and splits keys elements ($\mathbf{K}$) and values elements ($\mathbf{V}$) by $\mathbf{K}= W_s^K W_d^K \mathbf{Y_{HV}}$ and $\mathbf{V}= W_s^V W_d^V \mathbf{Y_{HV}}$.  
Then, the cross-attention is formulated as:
\begin{equation}
{\mathbf{\hat{Y}_I}} = \mathbf{V}\otimes \text{Softmax}\left(\mathbf{Q} \otimes \mathbf{K} / \alpha_H \right) + \mathbf{Y_I}
\end{equation}
where $\alpha_H$ represents multi-head factor similar to the multi-head self-attention but divide the number of channels into heads and $\otimes$ denotes matrix multiplication.
Finally, $\mathbf{\hat{Y}_I}$ are passed through a multilayer perceptron (MLP) to produce the light-up intensity map $\mathbf{\tilde{Y}_{I}}$.

The cross-attention between the dual-branch is used, rather than using self-attention individually on each branch. One main reason is that the illumination intensity is inversely proportional to the image noise intensity. The low-light image may also contain high-illumination regions that require only minimal denoising and enhancement. Therefore, using the intensity features to guide the HV-branch in denoising can reduce noise and mitigate color distortion in extremely dark regions. For another reason, the noisy intensity information, after being denoised in the HV-branch, is transferred to the I-branch through cross-attention, resulting in effective content restoration. Therefore, cross‑attention plays a pivotal role in mitigating color distortion and restoring content in extremely dark regions.

\begin{figure}
    \centering
    \includegraphics[width=1\linewidth]{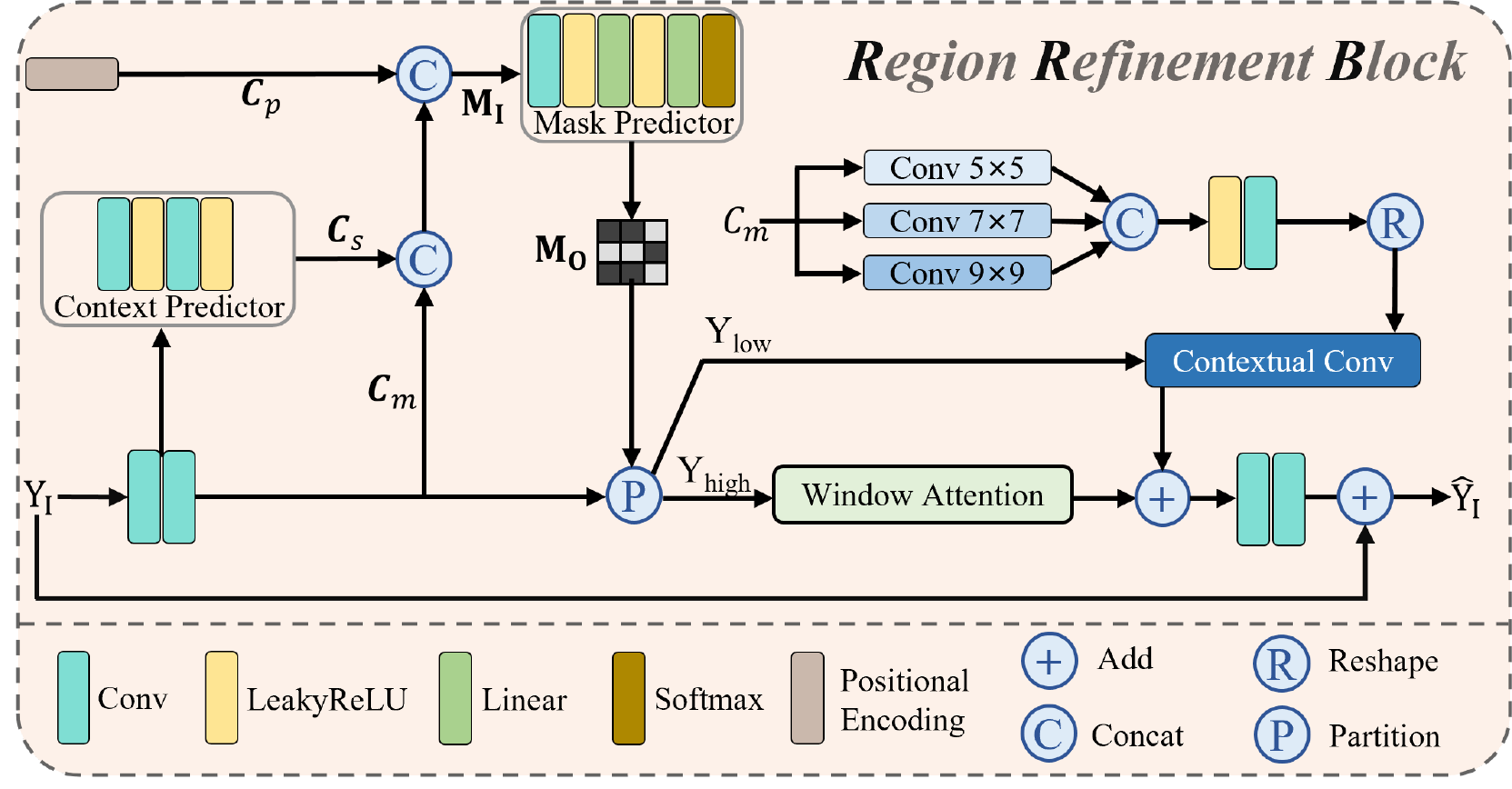}
    \caption{The architecture of Region Refinement Block (RRB). 
    The I‑branch feature map is convolved to extract mid‑level feature $\mathbf{C}_m$ and overview feature $\mathbf{C}_s$, which are combined with positional encoding $\mathbf{C}_p$ and fed to mask predictor that generates the partition mask.
    This mask splits $\mathbf{C}_m$ into information-scarce regions $\mathbf{Y}_\text{{high}}$ and information-rich regions $\mathbf{Y}_\text{{low}}$. For the $\mathbf{Y}_\text{{high}}$, window attention is implemented to aim in content restoration. For the $\mathbf{Y}_\text{{low}}$, a contextual multi-scale convolution strategy is employed to capture multi-scale context. The resulting feature maps are fused to predict dynamic small-kernel weights, which are then applied to the feature map to enhance details.}
    \label{fig:RRB}
\end{figure}

\textit{RRB.}
To respectively restore the severe degradation of content and details caused by uneven brightness in low-light images, we design the Region Refinement Block (RRB), which is inserted in the I-branch as a feature pre-processor, enabling the processing of different regions, as shown in Fig. \ref{fig:RRB}.
The RRB consists of two complementary branches: an attention‑based Transformer branch that leverages modeling of long-range dependence to restore content in regions where information is nearly lost (information-scarce regions), and a convolution‑based branch that employs local feature aggregation to enhance details in regions with sufficient cues (information-rich regions).

Given an input feature map $\mathbf{Y_I} \in \mathbb{R}^{H \times W \times C}$, we first extract a mid-level feature map $\mathbf{C}_m \in \mathbb{R}^{H \times W \times C}$ via convolutions and a semantically meaningful but low-quality overview feature map $\mathbf{C}_s \in \mathbb{R}^{H \times W \times 2}$ via convolutions followed by activation functions.
$\mathbf{C}_m$, $\mathbf{C}_s$ and the linear positional encoding $\mathbf{C}_p \in \mathbb{R}^{H \times W \times 2}$, which is generated from the feature and window dimensions to encode each patch’s spatial location, are concatenated along the channel axis to form the mask predictor input  $\mathbf{M_I} \in \mathbb{R}^{H \times W \times (C +4)}$.
Passing $\mathbf{M_I}$ through predictor yields partition mask $\mathbf{M_O}$.
We then split the $\mathbf{C}_m$ into information-scarce regions $\mathbf{Y}_\text{high}$ and information-rich regions $\mathbf{Y}_\text{low}$ by $\mathbf{M_O}$:
\begin{equation}
\mathbf{Y}_\text{high} = \mathbf{M_O} \odot \mathbf{C}_m, \quad \mathbf{Y}_\text{low} = (1 - \mathbf{M_O}) \odot \mathbf{C}_m
\end{equation}
where $\odot$ is the element-wise multiplication.

For the $\mathbf{Y}_{\text{low}}$, we propose a contextual multi-scale convolution strategy to establish global contextual links across the image, which captures long-range dependencies while preserving inherent local inductive biases. 
Then, a learnable module maps global context information into small-kernel convolution weights for feature extraction. As convolutional kernel slides over the feature map, each spatial token is adaptively modulated by global information, substantially enhancing the processing of information-rich regions features.

The strategy first captures multi-scale context by applying three parallel depthwise convolutions with kernel sizes $5$, $7$, and $9$.
We then fuse resulting feature maps by concatenation and a $1 \times 1$ projection to form a feature $\mathbf{F}$, from which we predict dynamic small-kernel weights via convolutions and GroupNorm.
Finally, each spatial location $(h, w)$ of $\mathbf{F}$ is convolved with \textit{contextual conv}, producing the output feature at channel $c$ and spatial location $(h,w)$:
\begin{equation}
\mathbf{S}_{[c, h, w]} = \sum_{i=0}^2 \sum_{j=0}^2 w_{ij}(h, w) \cdot \mathbf{F}_{[c, h + i -1, w + j -1]}
\end{equation}
Where $w_{ij}(\cdot)$ denotes the learnable weight at the spatial position $(h,w)$ corresponding to the relative offset $(i-1,j-1)$ and the $\mathbf{F}_{[c, h + i -1, w + j -1]}$ represents the fused input feature at the location shifted by $(i-1,j-1)$.
This contextual multi-scale convolution strategy effectively improves brightness and enhances details while allowing per-pixel adaptation via the learned dynamic small-kernel weights.

For the $\mathbf{Y}_{\text{high}}$, we implement the window attention mechanism~\cite{liu2021swin,wang2024camixersr} to capture remote information for better restoring content in information-scarce regions.
Finally, we fuse processed regions information via convolutions and the residual connection, thus generating $\hat{\mathbf{Y}}_\mathrm{I}$.

The RRB enables each decoder layer of the I-branch to dynamically restore content and enhance details in information-scarce regions and information-rich regions, effectively optimizing brightness adjustment.
\subsection{Feature Decoding}
\label{sec:feature_decoding}
In the feature decoding stage, we perform a Perceptual-inverse HVI Transformation (PHVIT), converting light-up HVI map back to the sRGB color space, which is a subjective mapping while allowing independent adjustment of the image’s saturation and brightness.

PHVIT sets $\hat{h}$ and $\hat{v}$ as intermediate variables as $\hat{h}=\frac{ \mathbf{\hat{H}}}{\mathbf{C}_k+\mathcal{\varepsilon}}$,$\hat{v}=\frac{ \mathbf{\hat{V}}}{\mathbf{C}_k+\mathcal{\varepsilon}}$, where $\mathcal{\varepsilon}=1\times10^{-8}$ is used to avoid gradient explosion. Then, we convert $\hat{h}$ and $\hat{v}$ to HSV color space. The Hue ($\mathbf{H}$), Saturation ($\mathbf{S}$) and Value ($\mathbf{V}$) map can be estimated as:
\begin{equation}
\begin{split}
    \mathbf{H} &= \arctan(\frac{\hat{v}}{\hat{h}})\mod 1,\\
    \mathbf{S} &= \alpha_{S}\sqrt{\hat{h}^{2}+\hat{v}^{2}},\\
    \mathbf{V} &= \alpha_{I}\hat{\mathbf{I}}_{\mathbf{I}},
\end{split}
\end{equation}
where $\alpha_{S},\alpha_{I}$ are the customizing linear parameters to change the color saturation and brightness of the image. Finally, we will obtain the sRGB image with the HSV image \cite{Foley1982FundamentalsOI}.

\subsection{Loss Function}
For training HVI-CIDNet+, we introduced complementary loss functions in both the sRGB and HVI color spaces to furnish more exhaustive supervision. 
In the sRGB domain, the loss penalizes the discrepancy between the enhanced output $\mathbf{\hat{I}}$ and the Ground Truth image $\mathbf{I}$, whereas in the HVI domain it penalizes the difference between the enhanced HVI map $\mathbf{\hat{I}_{HVI}}$ and the Ground Truth HVI map $\mathbf{I_{HVI}}$.
By supervising the network simultaneously in these two perceptually meaningful spaces, the model is guided to learn richer enhancement characteristics and patterns, yielding superior final results. The combined loss is defined as follows:
\begin{equation}
L= \lambda\cdot l(\mathbf{\hat{I}_{HVI}},\mathbf{I_{HVI}}) + l(\mathbf{\hat{I}},\mathbf{I}),
\end{equation}
where $\lambda$ is a weighting hyperparameter to balance the losses in the two different color spaces. This helps achieve not only more closely with the probabilistic distribution of sRGB in the HVI color space, especially the red and black ones, due to the optimized $k$ and $\mathbf{C}_k$, but also the inheritance of pixel-level structure detail in the sRGB color space.

\begin{table*}[!t]
    \centering
    \renewcommand{\arraystretch}{1.}
    \caption{Quantitative results of PSNR/SSIM$\uparrow$ and LPIPS$\downarrow$ on the LOL (v1 and v2) datasets. Due to the limited number of test set in LOLv1, we use GT mean method during testing to minimize errors. The FLOPs is tested on a single $256\times256$ image. The best performance is in \textcolor{red}{red} color and the second best is in \textcolor{cyan}{cyan} color.}
    \vspace{-1.8mm}
    \resizebox{\textwidth}{!}{
    \begin{tabular}{c|c|cc|ccc|ccc|ccc}
        \hline
        \multirow{2}{*}{\textbf{Methods}}&\multirow{2}{*}{\textbf{Color Model}}&\multicolumn{2}{c|}{\textbf{Complexity}}&\multicolumn{3}{c|}{\textbf{LOLv1}} & \multicolumn{3}{c|}{\textbf{LOLv2-Real}} & \multicolumn{3}{c}{\textbf{LOLv2-Synthetic}}\\
        ~&~&Params/M&FLOPs/G&PSNR$\uparrow$&SSIM$\uparrow$&LPIPS$\downarrow$&PSNR$\uparrow$&SSIM$\uparrow$&LPIPS$\downarrow$&PSNR$\uparrow$&SSIM$\uparrow$&LPIPS$\downarrow$\\
        \hline
        RetinexNet \cite{RetinexNet}& Retinex& 0.84& 584.47 &	18.915 &	0.427 & 0.470&	16.097 &	0.401 	&0.543 &	17.137 &	0.762 &	0.255
\\
        KinD \cite{KinD}& Retinex & 8.02& 34.99 &23.018& 	0.843& 0.156&	17.544& 	0.669 &	0.375	&	18.320 	&0.796 &0.252
\\
         ZeroDCE \cite{Zero-DCE}& RGB&0.08&4.83& 	21.880 &	0.640&0.335& 16.059 &	0.580 & 0.313&	17.712& 	0.815 &	0.169
\\
         ZeroDCE++ \cite{li2021learning}& RGB&0.01 &0.02 &18.516 &0.434&0.360&17.230 &0.412 &0.319&17.577&0.812 &0.187
\\
        LLFlow \cite{LLFlow}& RGB&17.42 & 358.4 &24.998 &	0.871&	0.117&17.433 &	0.831 & 0.176&	24.807 &	0.919 &	0.067
\\
         EnlightenGAN \cite{EnGAN}& RGB& 114.35 & 61.01 &  20.003 & 0.691&0.317 &18.230 & 0.617 &0.309 & 16.570& 0.734&0.220
\\

         SNR-Aware \cite{SNR-Aware}& RGB& 4.01 & 26.35 &26.716 	&0.851 	&0.152 &21.480 &	0.849 &0.163	&	24.140 &	0.928 & 0.056
\\
        Bread \cite{Bread}& YCbCr&2.02 & 19.85 &25.299 	&0.847 	&0.155 &20.830 &	0.847 &0.174	&	17.630 &	0.919 & 0.091
\\
        LYT-Net \cite{brateanu2025lyt}& YCbCr&0.05 &3.49  &26.583 &0.835 &0.130 &20.965 &0.841 &0.139 &23.496 &0.916 &0.085 
\\
        PairLIE \cite{PairLIE}& Retinex& 0.33 & 20.81 &	23.526 &	0.755& 0.248&	19.885 &	0.778 & 0.317&	19.074	&0.794&	0.230
\\
        LLFormer \cite{LLFormer}& RGB& 24.55&22.52&25.758&0.823&0.167&20.056&0.792& 0.211&24.038&0.909&0.066
\\
         RetinexFormer \cite{RetinexFormer}& Retinex& 1.53 & 15.85 & 	27.140 & 	0.850  & 0.129&	22.794  &	0.840  & 0.171&	25.670  &	0.930 &	0.059
\\
        GSAD \cite{hou2024global}& RGB& 17.36 & 442.02 & 	27.605& 	0.876  &0.092&	20.153  &	0.846  & 0.113&	24.472 &	0.929  &	0.051
\\
       QuadPrior \cite{wang2024zero}& Kubelka-Munk& 1252.75 & 1103.20 & 	22.849& 	0.800  & 0.201&	20.592  &	0.811  & 0.202&	16.108 &	0.758 &	0.114
\\
        CoLIE \cite{chobola2024fast}& HSV&0.13 &8.06 &20.396 	&0.478 	&0.373 &15.072 &0.501&0.322&14.291 &0.654 & 0.250
\\
        Zero-IG \cite{shi2024zero} & Retinex& 0.08 & 30.19 & 26.077 & 0.794 & 0.191&18.132& 0.746 & 0.248& 15.777 & 0.762 & 0.259
\\
        LightenDiff \cite{Jiang_2024_ECCV} & Retinex& 26.54 & 2257.42& 23.620 & 0.829 & 0.180&22.878& 0.855 & 0.166& 21.582 & 0.869 & 0.153
\\
        CIDNet \cite{yan2025hvi}& HVI& 1.88 & 7.57 & 	\color{cyan}{28.201} &	\color{cyan}{0.889} &\color{cyan}{0.079}	&\color{cyan}{24.111}& 	\color{cyan}{0.871} &\color{cyan}{0.108}  &	\textcolor{cyan}{25.705} &	\color{cyan}{0.942} & \color{cyan}{0.045}
\\
        \textbf{HVI-CIDNet+}&HVI
        &300.88  &23.54 	&\color{red}{28.846} &	\color{red}{0.894} &\color{red}{0.058}	&\color{red}{24.308}& 	\color{red}{0.873} &\color{red}{0.107}  &	\textcolor{red}{25.982} &	\color{red}{0.943} & \color{red}{0.040}
\\
         \hline
    \end{tabular}
    }
    \label{tab:table-LOL}
\end{table*}

\begin{table}[t]
\centering
\caption{Datasets summary on low-light image enhancement.}
\renewcommand{\arraystretch}{1.2}
\label{tab:dataset}
\resizebox{\linewidth}{!}{
\begin{tabular}{l|l|rr|c}
\cline{1-5}
\textbf{Dataset} & \textbf{Subsets} & \textbf{\#Train} & \textbf{\#Test} & \textbf{Resolutions ($H\times W$)} \\
\cline{1-5}
\multirow{3}{*}{\textbf{LOL}} & v1 \cite{RetinexNet} & 485 & 15 & $400\times600$ \\
& v2-Real \cite{LOLv2} & 689 & 100 & $400\times600$ \\
& v2-Synthetic \cite{LOLv2}& 900 & 100 & $384\times384$ \\
\cline{1-5}
\multirow{5}{*}{\makecell{\textbf{Unpaired} \\\textbf{Datasets}}} & DICM \cite{DICM} & 0 & 69 & Various \\
& LIME \cite{LIME} & 0 & 10 & Various \\
& MEF \cite{MEF} & 0 & 17 & Various \\
& NPE \cite{NPE} & 0 & 8 & Various \\
& VV \cite{VV} & 0 & 24 & Various \\
\cline{1-5}
\multirow{3}{*}{\textbf{SICE}} & Original \cite{SICE} & 4800 & 0 & Various \\
& Mix \cite{SICE-Mix} & 0 & 589 & $600\times900$ \\
& Grad \cite{SICE-Mix} & 0 & 589 & $600\times900$ \\
\cline{1-5}
\textbf{SID }\cite{SID}& Sony-Total-Dark & 2099 & 598 & $1424\times2128$ \\
\cline{1-5}
\end{tabular}
}
\end{table}

\section{Experiments}
\subsection{Datasets and Settings}
We evaluate our method on ten widely adopted LLIE benchmark datasets: LOLv1 \cite{RetinexNet}, LOLv2 \cite{LOLv2}, unpaired datasets (DICM \cite{DICM}, LIME \cite{LIME}, MEF \cite{MEF}, NPE \cite{NPE}, and VV \cite{VV}), SICE \cite{SICE}(including Mix and Grad test sets~\cite{SICE-Mix}), and SID (Sony-Total-Dark) \cite{SID}. All the datasets used in the paper are summarized in Table~\ref{tab:dataset}.

\textit{LOL.}
The LOL dataset consists of versions v1 and v2. LOL-v2 offers distinct subsets for real data and synthetic data.


\textit{SICE.}
The original SICE dataset \cite{SICE} covers multiple scenes and high-resolution multi-exposure sequences, which are divided into training, validation, and test sets in a 7: 1: 2 ratio.

\textit{Sony-Total-Dark.}
The dataset is a customized subset of the SID dataset \cite{SID}, specially designed to increase the challenge of the LLIE tasks. To achieve this, we transform the original raw-format images into sRGB format without applying gamma correction, resulting in images that exhibit extreme darkness.

\textit{Experiment Settings.}
We implement HVI-CIDNet+ using PyTorch. The HVI-CIDNet+ is trained with the Adam \cite{Adam} optimizer (\textit{$\beta_{1}$} = 0.9 and \textit{$\beta_{2}$} = 0.999) using a single NVIDIA 4090 GPU. The learning rate is initially set to $1 \times 10^{-4}$ and then steadily decreased to $1 \times 10^{-7}$ by the cosine annealing scheme \cite{sgdr} during the training process.
For the LOL benchmarks, we crop $384\times384$ patches from paired low-light and normal-light images. Training continues for $1,000$ epochs with a batch size of $4$, using random rotations and flips for augmentation. Regarding SICE and SID, we extract $256\times256$ patches and train for $500$ epochs with a batch size of $8$, applying the same enhancement strategy.

\textit{Evaluation Metrics.}
For the paired datasets, We evaluate the distortion using Peak Signal-to-Noise Ratio (PSNR) and Structural Similarity (SSIM) \cite{SSIM}. In addition, to assess the perceptual quality of restored images, we use the Learned Perceptual Image Patch Similarity (LPIPS) \cite{LPIPS}, using AlexNet \cite{Alex} as the reference network. For the unpaired datasets, we perceptually evaluate individual restored images using BRISQUE \cite{BRISQUE} and NIQE \cite{NIQE}.
\begin{table*}[t]
    \centering
    \renewcommand{\arraystretch}{1.}
    \caption{Results of applying HVI transformation as a plug-in to various LLIE methods on LOLv2-Real dataset. 
    \textcolor{red}{Values} in brackets represent the absolute improved performance gain. The best PSNR/SSIM$\uparrow$ and LPIPS$\downarrow$ are in \textbf{bolded}.
    } 
    \vspace{-1.8mm}
    \resizebox{\linewidth}{!}{
    \begin{tabular}{c|c|c|c|c|c|c|c|c}
    \hline
         Methods& 
         FourLLIE \cite{wang2023fourllie}&
         LEDNet \cite{LEDNet}&
         SNR-Aware \cite{SNR-Aware}& 
         LLFormer \cite{LLFormer}& 
         GSAD \cite{GSAD}& 
         DiffLight \cite{feng2024difflight}&
        \textbf{CIDNet} \cite{yan2025hvi}&
        \textbf{HVI-CIDNet+}\\
    \hline
         PSNR$\uparrow$&
         22.730(\textcolor{red}{+0.381})&
         23.394(\textcolor{red}{+3.456})&
         22.251(\textcolor{red}{+0.771})&
         22.671(\textcolor{red}{+2.615})&		
         23.715(\textcolor{red}{+3.562})&	
         23.969(\textcolor{red}{+1.364})&
         24.111&
         \textbf{24.308}
\\
         SSIM$\uparrow$&
         0.856(\textcolor{red}{+0.009})&
         0.837(\textcolor{red}{+0.010})&
         0.840(-0.009)&	
         0.852(\textcolor{red}{+0.060})&		
         \textbf{0.876}(\textcolor{red}{+0.030})&	
         0.859(\textcolor{red}{+0.003})&
         0.871&
         0.873

\\
        LPIPS$\downarrow$&
        0.125(+0.011)&
        0.115(\textcolor{red}{-0.005})&	
        0.117(\textcolor{red}{-0.054})&	
        0.117(\textcolor{red}{-0.094})&	
        \textbf{0.103}(\textcolor{red}{-0.010})&	
        0.109(\textcolor{red}{-0.012})&
        0.108&
        0.107
\\

        Model Type&
        CNN&
        CNN&
        Transformer&
        Transformer&		
        Diffusion&
        CNN+Diffusion&
        Transformer&
        Transformer
\\
    \hline
    \end{tabular}
    }
    \label{tab:HVI}
\end{table*}

\begin{table}
      \centering
        \renewcommand{\arraystretch}{1.}
        \caption{Quantitative result on SICE, Sont-Total-Dark, and the five unpaired datasets (DICM \cite{DICM}, LIME \cite{LIME}, MEF \cite{MEF}, NPE \cite{NPE}, and VV \cite{VV}). The top-ranking score is in \textcolor{red}{red} color.}
        \vspace{-1.8mm}
        \resizebox{\linewidth}{!}{
        \begin{tabular}{c|cc|cc|cc}
        \hline
        \multirow{2}{*}{\textbf{Methods}}&
         \multicolumn{2}{c|}{\textbf{SICE}}& 
         \multicolumn{2}{c|}{\textbf{Sony-Total-Dark}} & 
         \multicolumn{2}{c}{\textbf{Unpaired}}\\

        ~&
            PSNR$\uparrow$&	SSIM$\uparrow$& 
            PSNR$\uparrow$&	SSIM$\uparrow$&
            BRIS$\downarrow$& NIQE$\downarrow$\\
            
            \hline

            RetinexNet \cite{RetinexNet}&
            12.424& 0.613&
            15.695& 0.395& 
            23.286& 4.558
\\
            ZeroDCE \cite{Zero-DCE}&
            12.452& 0.639& 
            14.087&	0.090& 
            26.343&	4.763
\\
            URetinexNet \cite{URetinexNet}&
            10.899& 0.605& 
            15.519& 0.323&
            26.359& 3.829
\\
            RUAS \cite{RUAS}&
            8.656& 	0.494& 
            12.622& 0.081&  
            36.372& 4.800 
\\
            LLFlow \cite{LLFlow}&
            12.737&	0.617& 
            16.226&	0.367& 
            28.087&	4.221 
\\
            CIDNet \cite{yan2025hvi}&
            13.435&	0.642&
            22.904&	0.676&
            23.521&	3.523

\\          \textbf{HVI-CIDNet+}&
            \color{red}{14.450}&	\color{red}{0.654}&
            \color{red}{23.482}&	\color{red}{0.691}&
            \color{red}{22.675}&	\color{red}{3.467}
\\
            \hline
        \end{tabular}
        }
        \label{tab:SID}
\end{table}

\subsection{Main Results}
In this section, we evaluate the performance of HVI-CIDNet+ on various benchmark datasets. We report results on paired and unpaired LLIE tasks, as well as the effectiveness of the proposed HVI color space.

\begin{figure*}[!t]
\centering
\begin{minipage}[t]{0.19\linewidth}
    \centering
    \vspace{1pt}
    \centerline{\includegraphics[width=\textwidth]{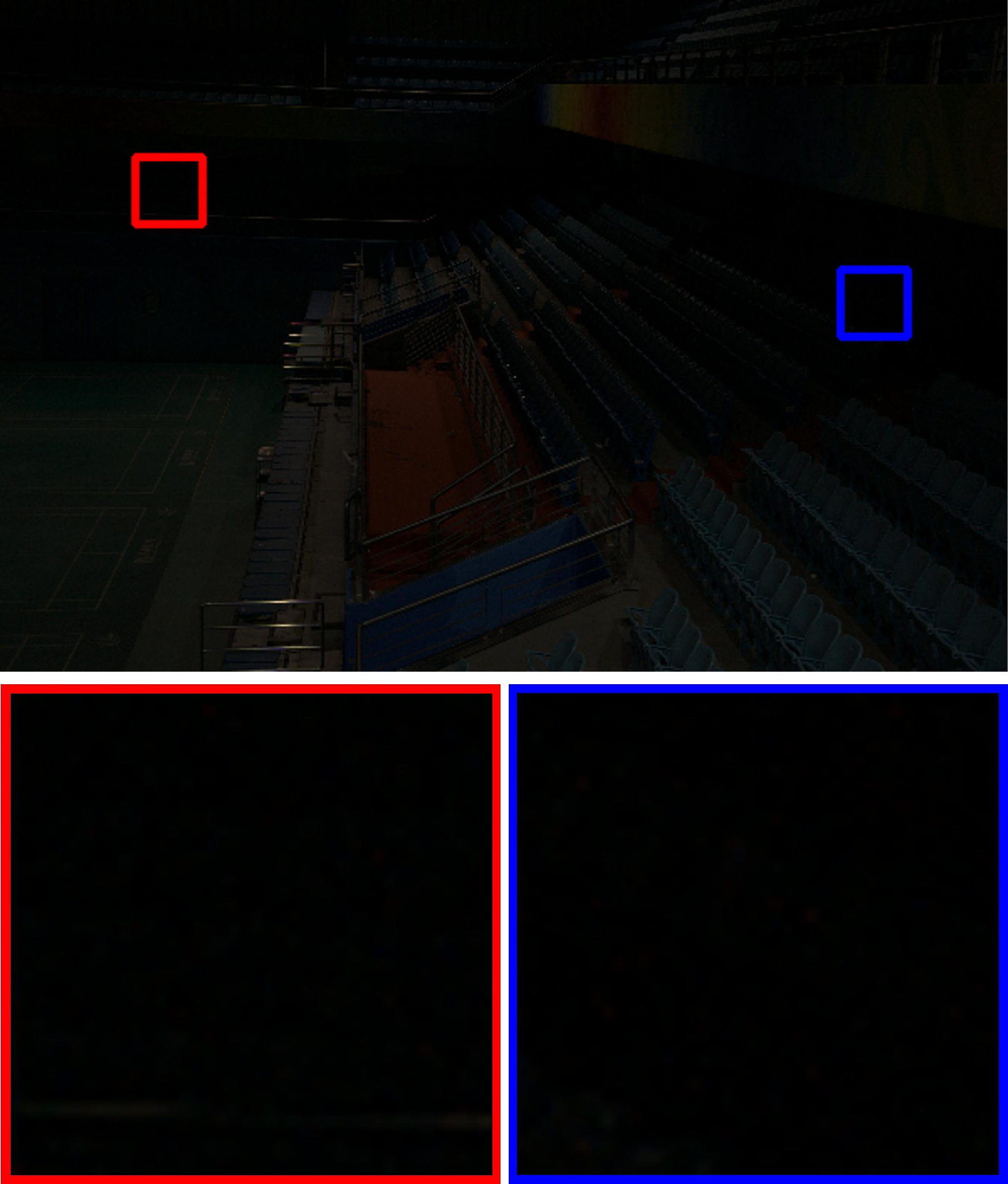}}
    \centerline{\small 8.877/0.081}
    \centerline{\small Input}
    \vspace{1pt}
    \centerline{\includegraphics[width=\textwidth]{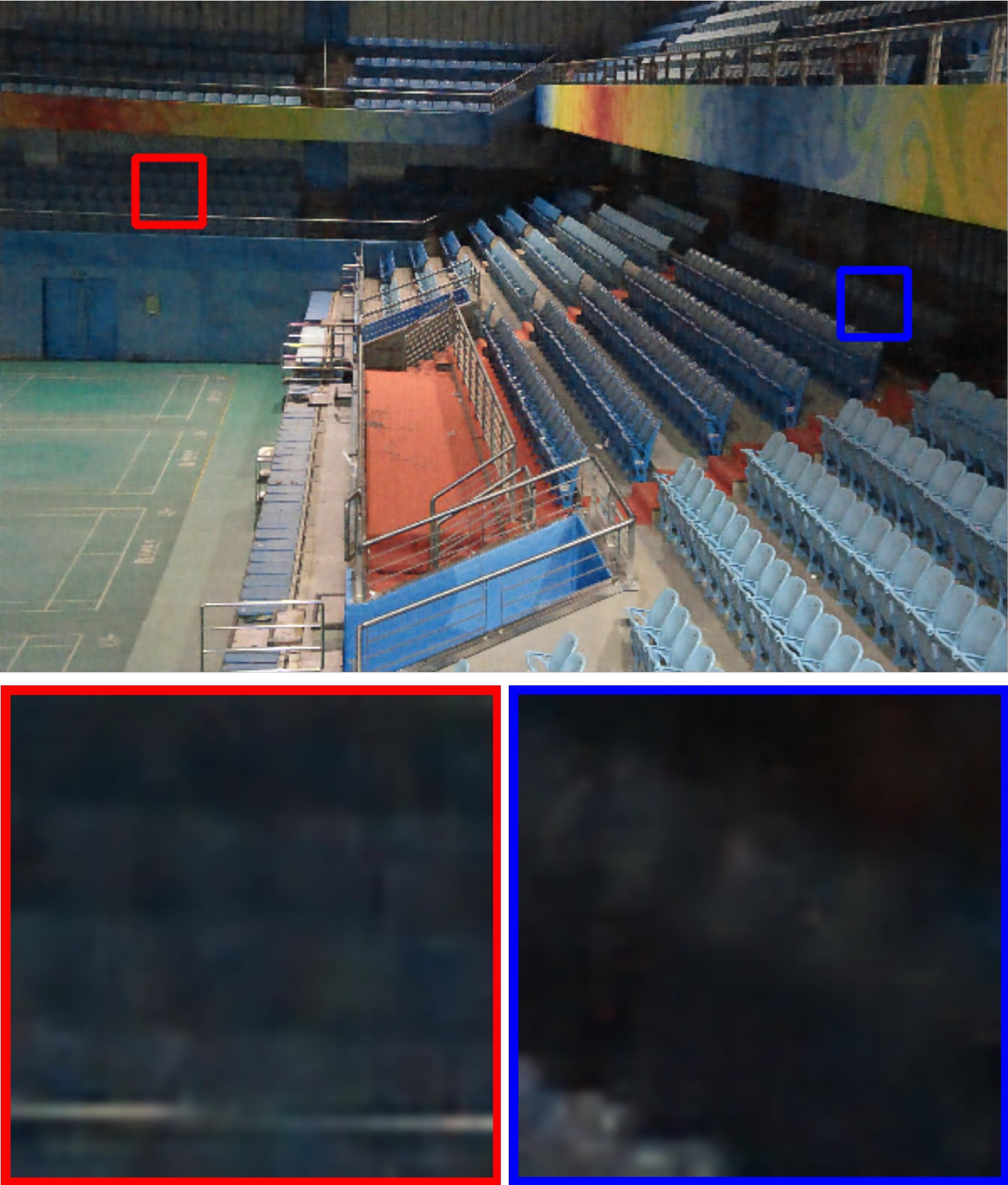}}
    \centerline{\small 24.619/0.786}
    \centerline{\small RetFormer \cite{RetinexFormer}}
    \vspace{4pt}
    \centerline{\includegraphics[width=\textwidth]{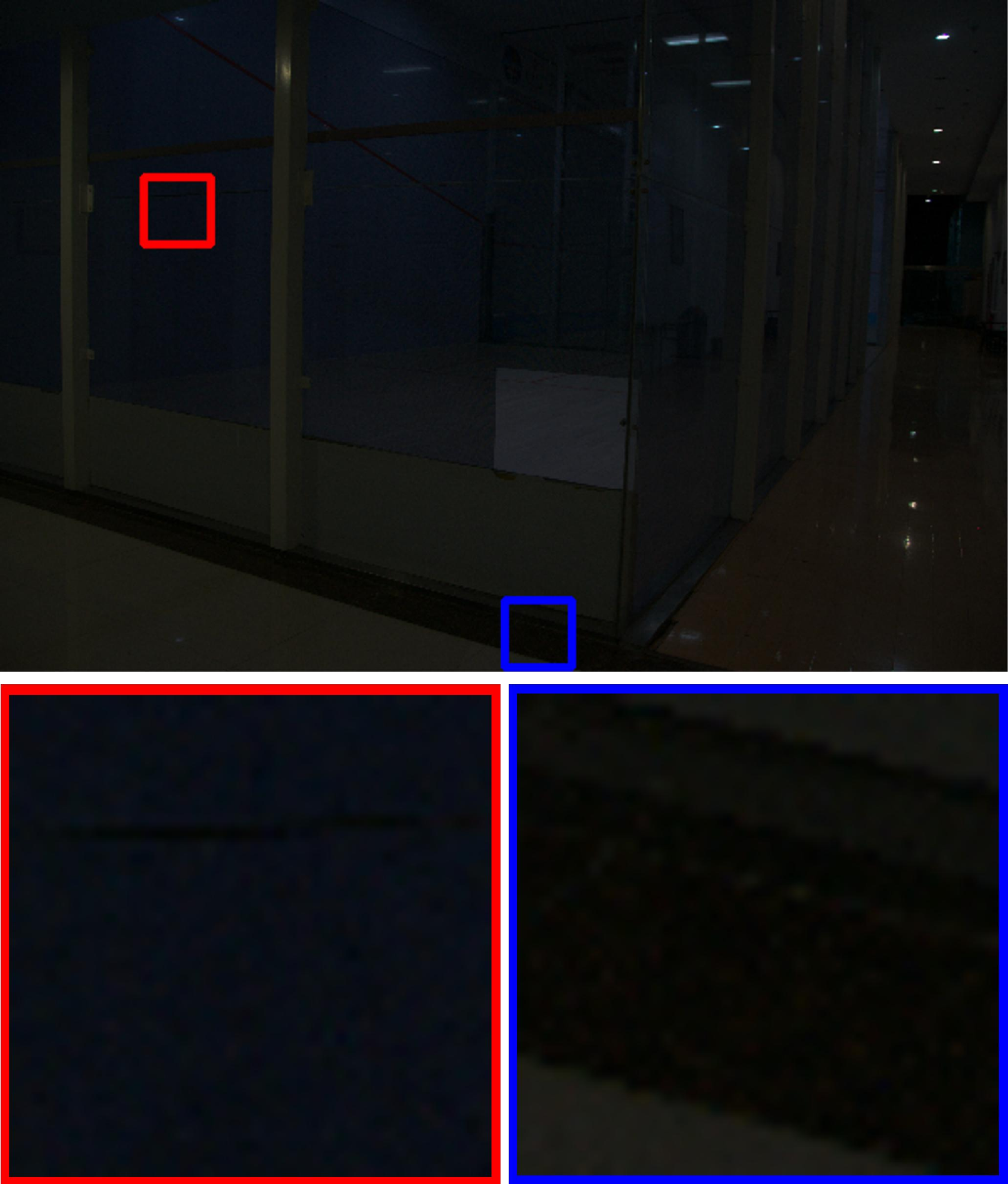}}
    \centerline{\small 10.032/0.282}
    \centerline{\small Input}
    \vspace{1pt}
    \centerline{\includegraphics[width=\textwidth]{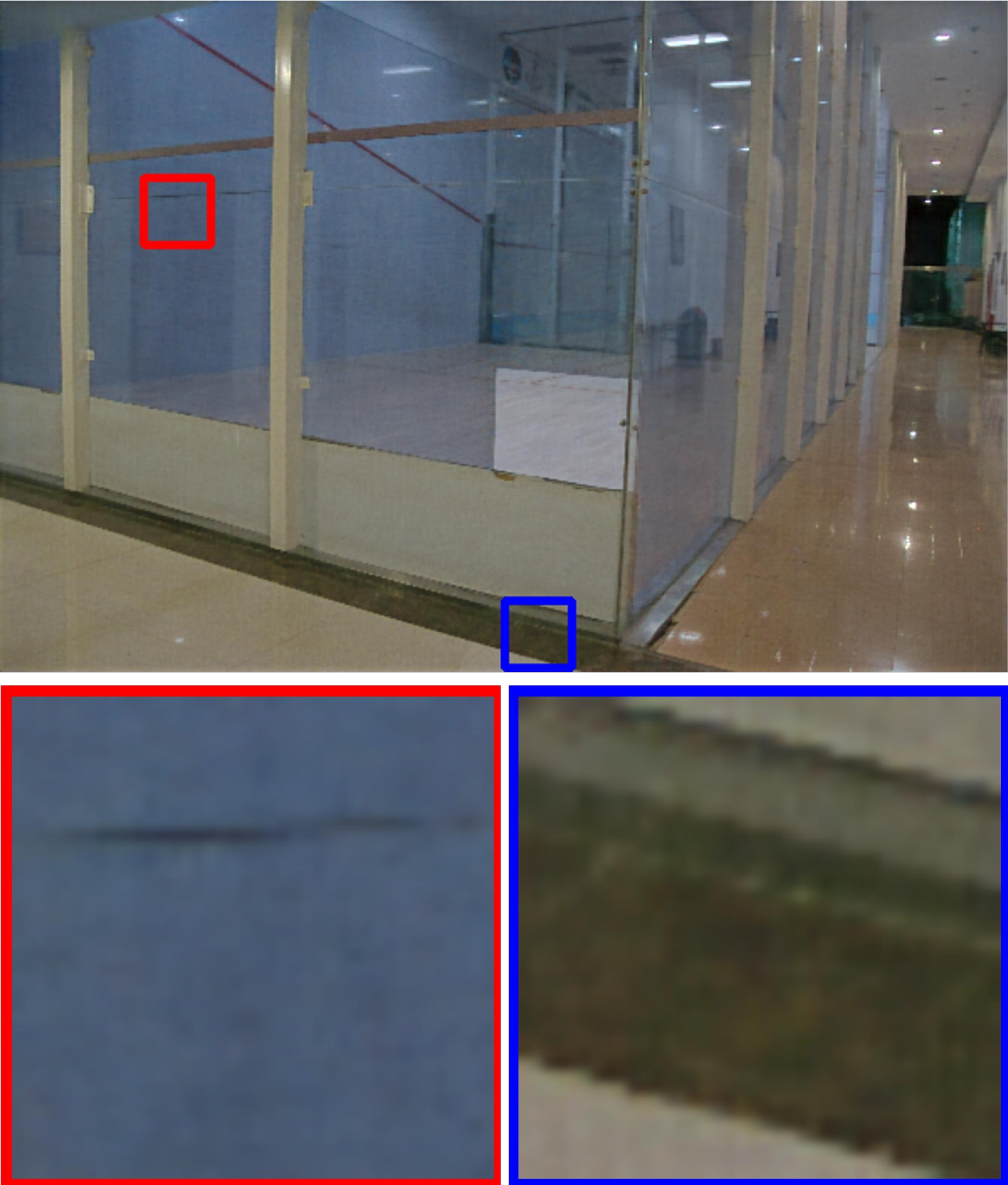}}
    \centerline{\small 28.741/0.882}
    \centerline{\small RetFormer \cite{RetinexFormer}}
    
\end{minipage}
\hfill
\begin{minipage}[t]{0.19\linewidth}
    \centering
    \vspace{1pt}
    \centerline{\includegraphics[width=\textwidth]{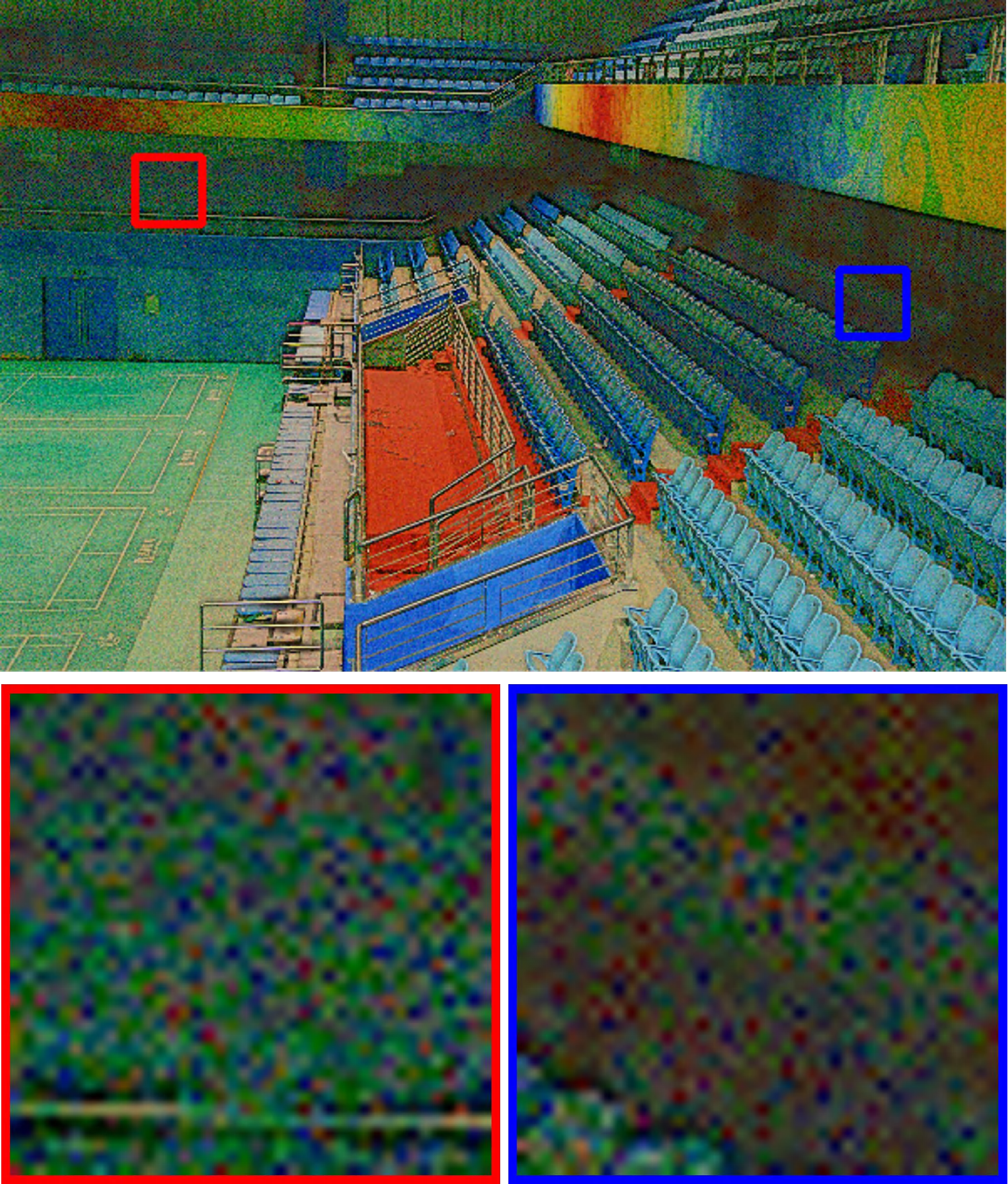}}
    \centerline{\small 18.430/0.421}
    \centerline{\small RetinexNet \cite{RetinexNet}}
    \vspace{1pt}
    \centerline{\includegraphics[width=\textwidth]{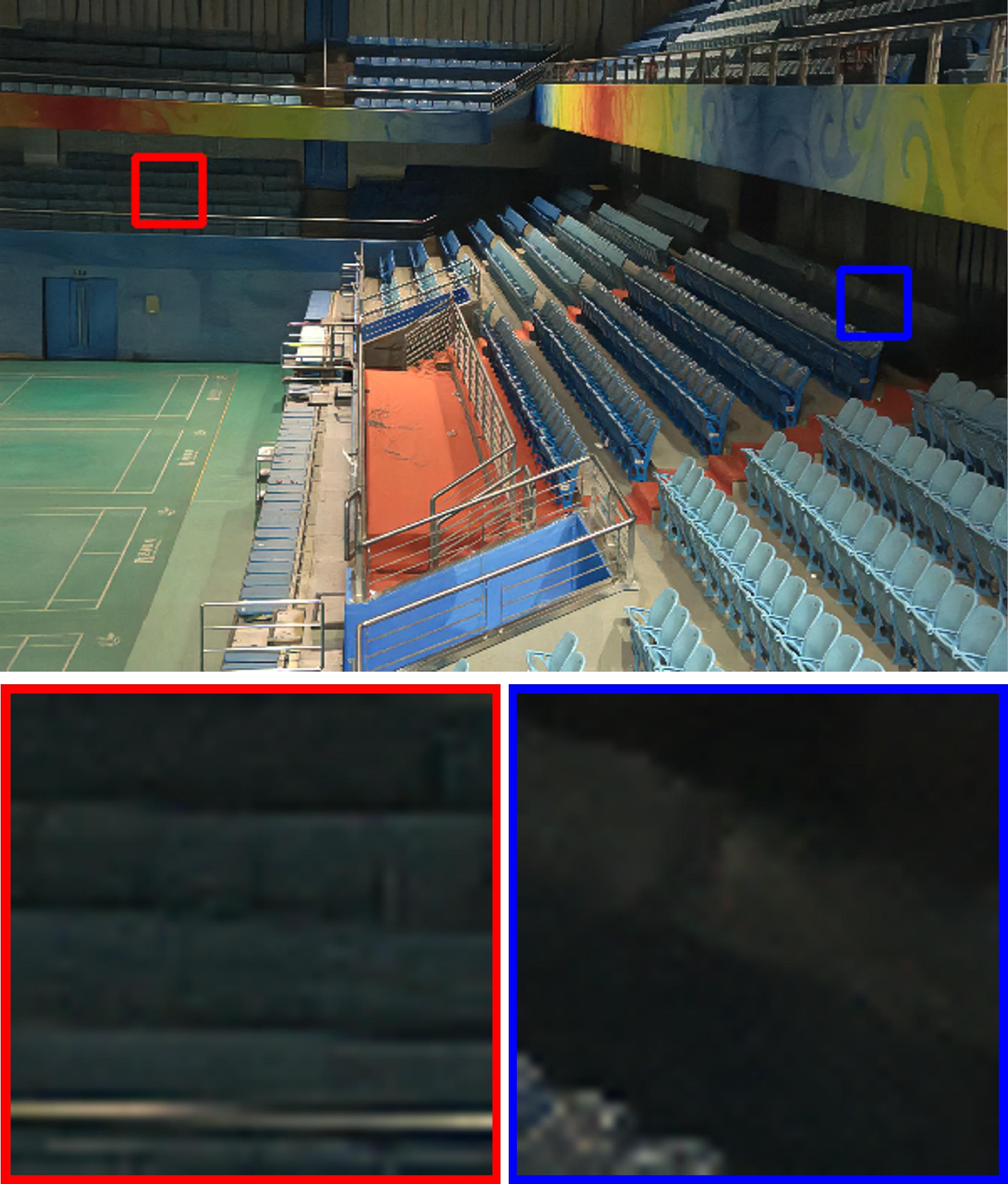}}
    \centerline{\small 24.915/0.813}
    \centerline{\small GSAD \cite{GSAD}}
    \vspace{4pt}
    \centerline{\includegraphics[width=\textwidth]{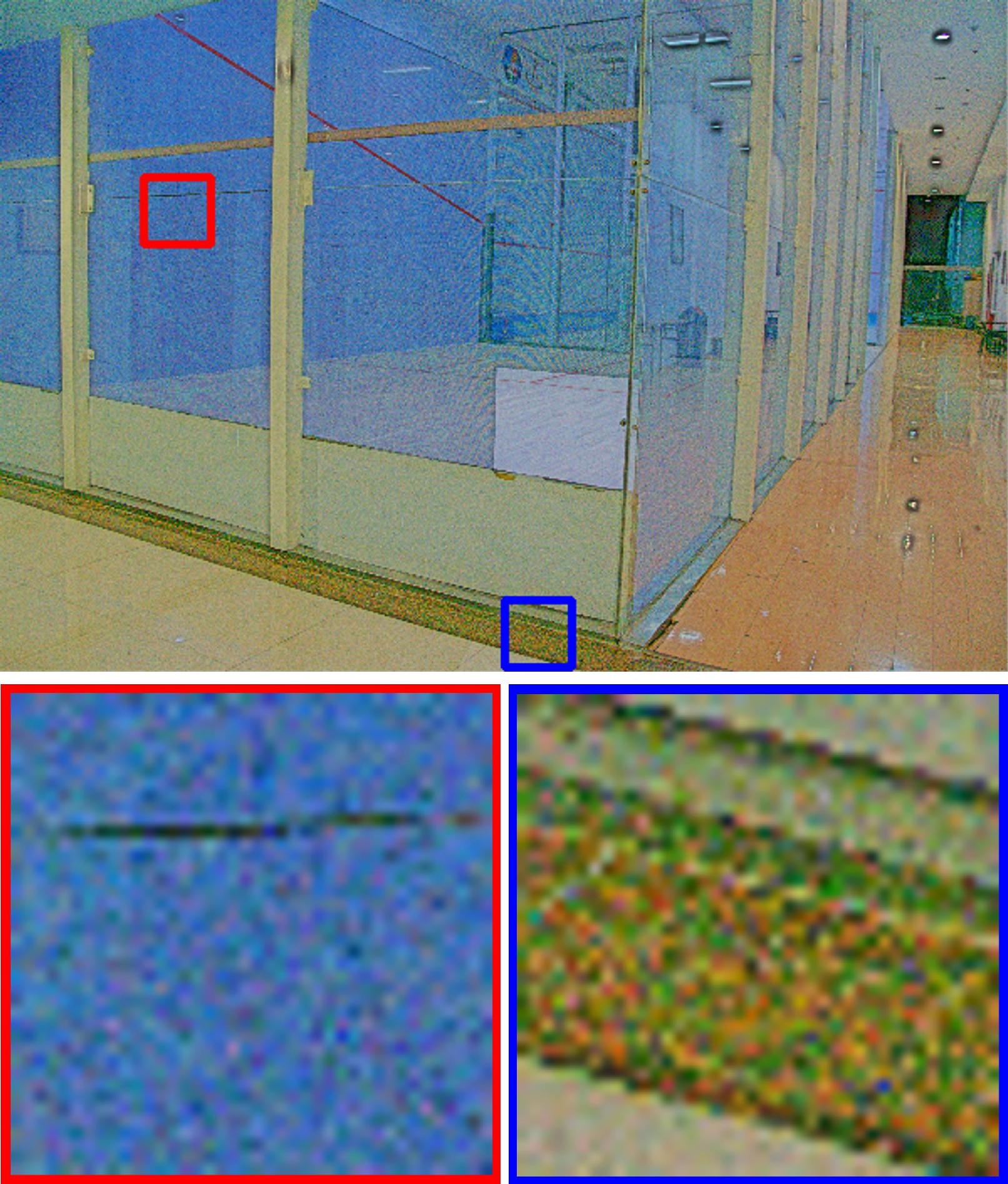}}
    \centerline{\small 16.398/0.382}
    \centerline{\small RetinexNet \cite{RetinexNet}}
    \vspace{1pt}
    \centerline{\includegraphics[width=\textwidth]{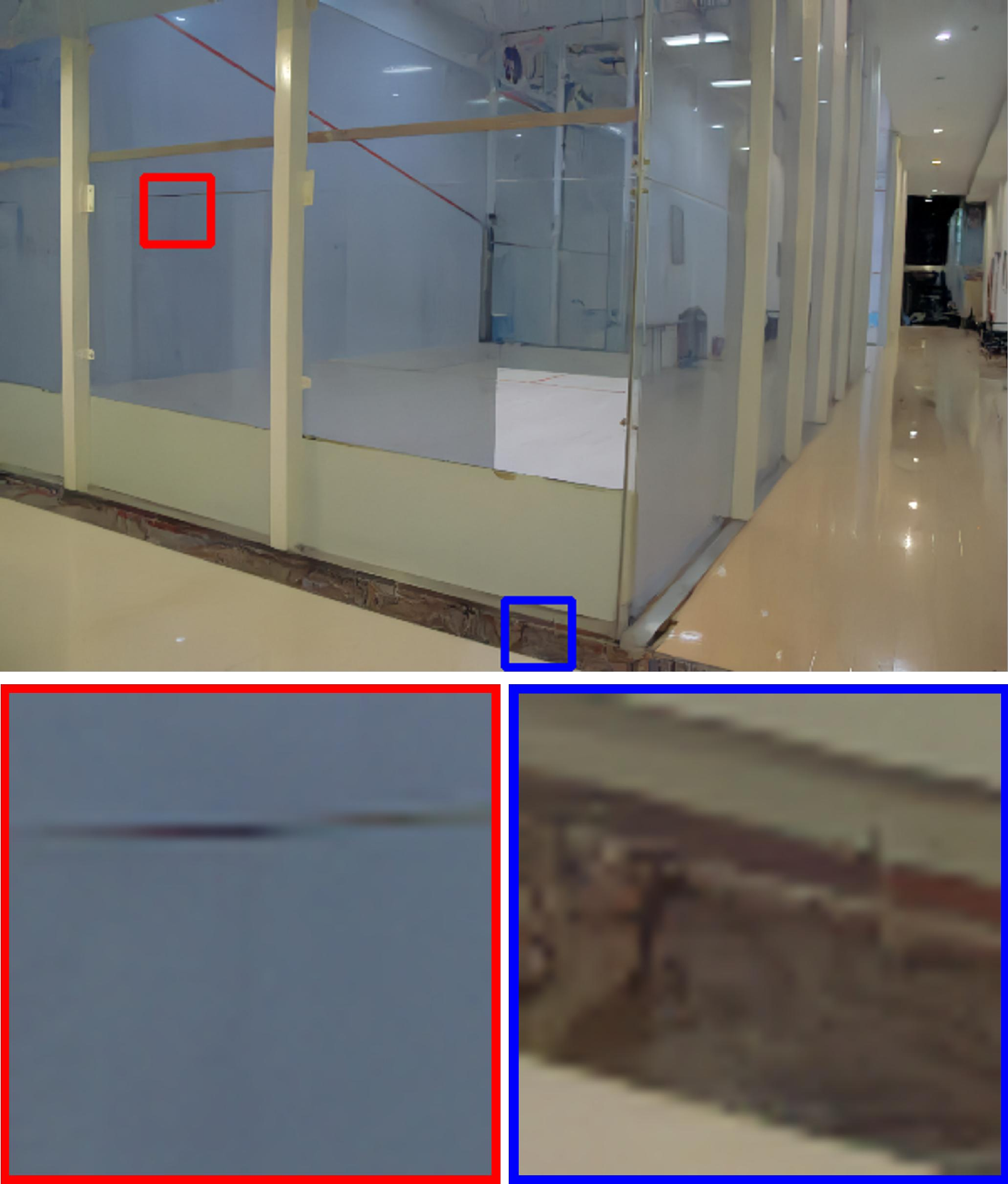}}
    \centerline{\small 20.223/0.857}
    \centerline{\small GSAD \cite{GSAD}}
\end{minipage}
\hfill
\begin{minipage}[t]{0.19\linewidth}
    \centering
    \vspace{1pt}
    \centerline{\includegraphics[width=\textwidth]{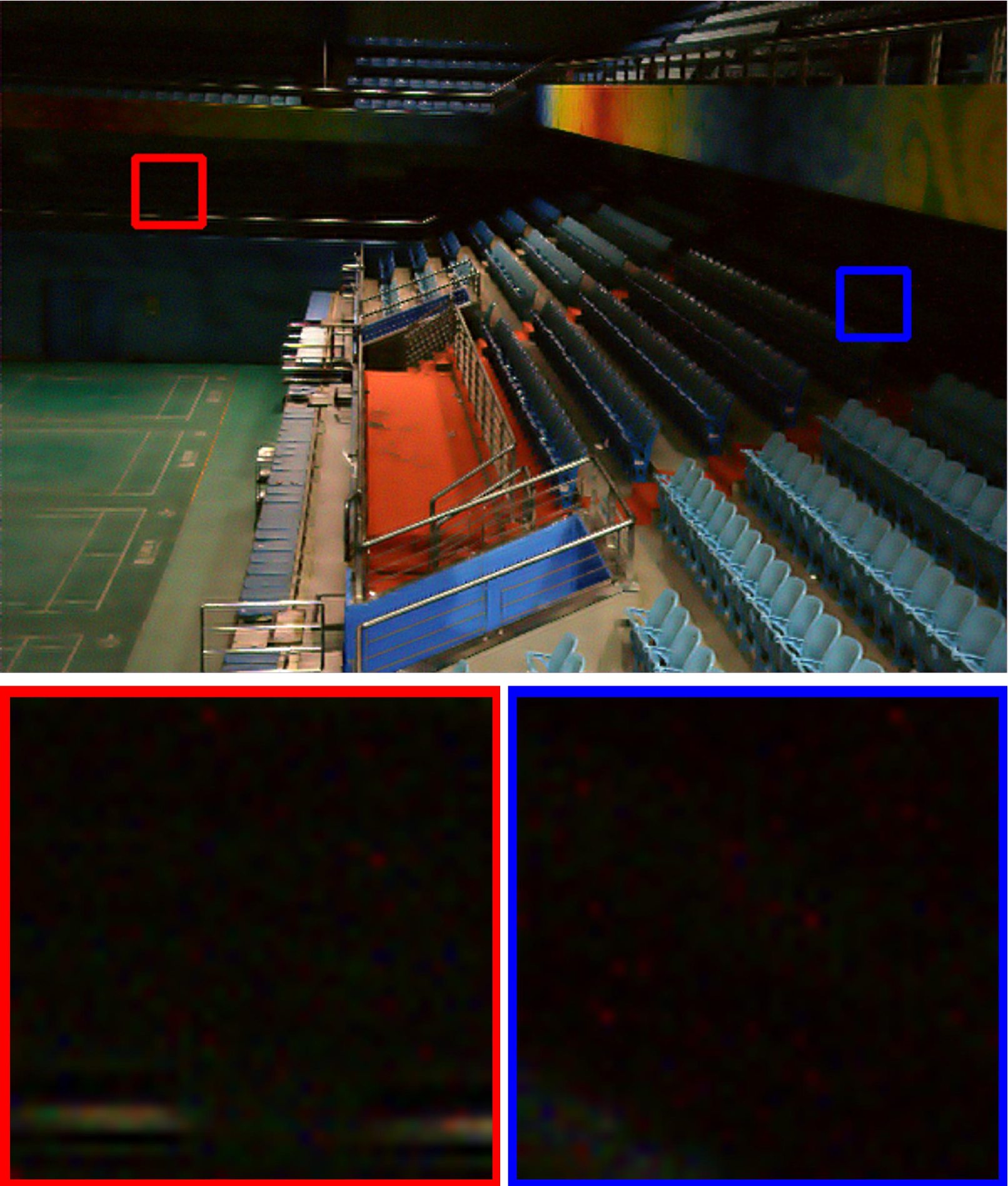}}
    \centerline{\small 14.768/0.405}
    \centerline{\small RUAS \cite{RUAS}}
    \vspace{1pt}
    \centerline{\includegraphics[width=\textwidth]{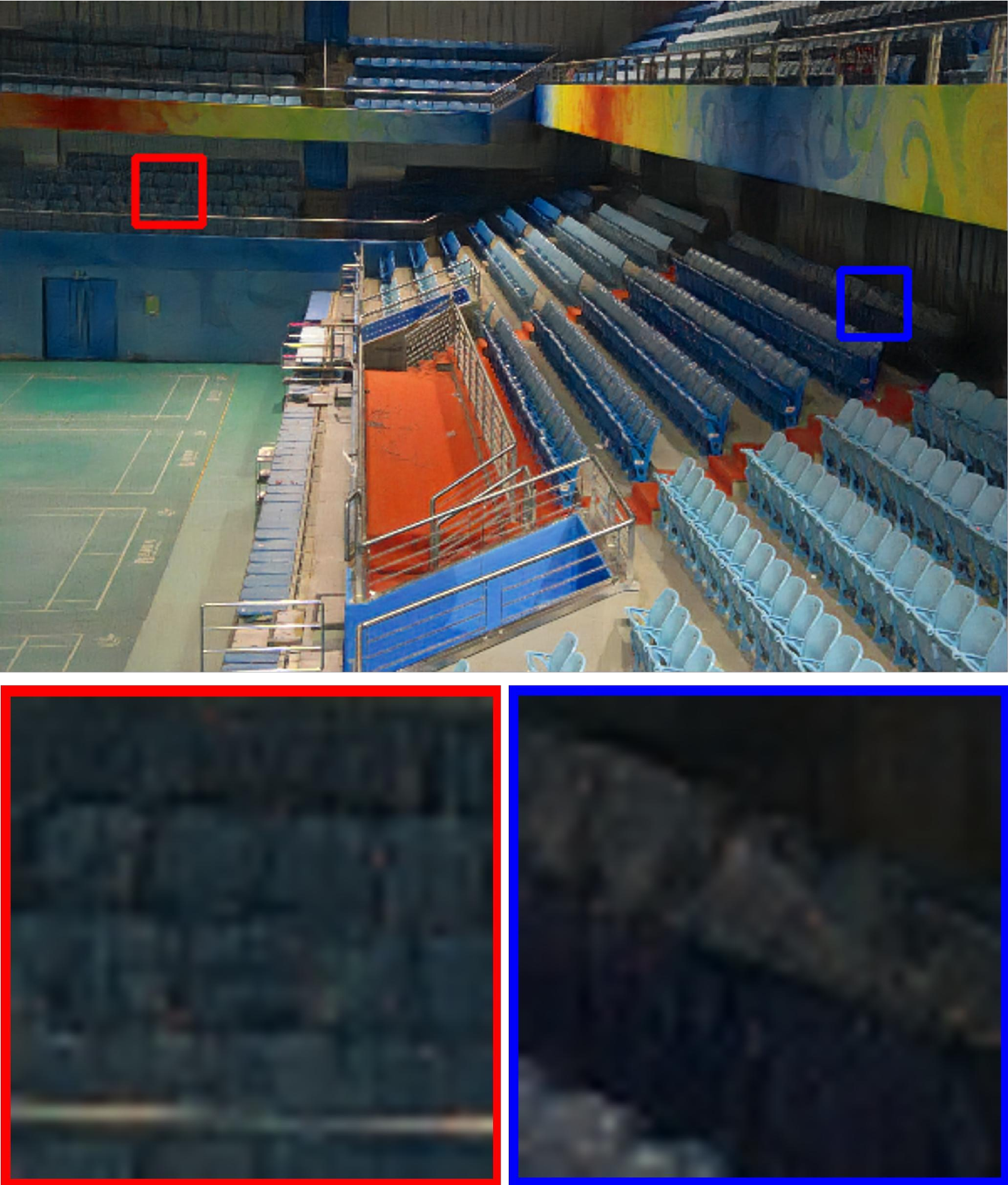}}
    \centerline{\small 25.576/0.803}
    \centerline{\small CIDNet \cite{yan2025hvi}}
    \vspace{4pt}
    \centerline{\includegraphics[width=\textwidth]{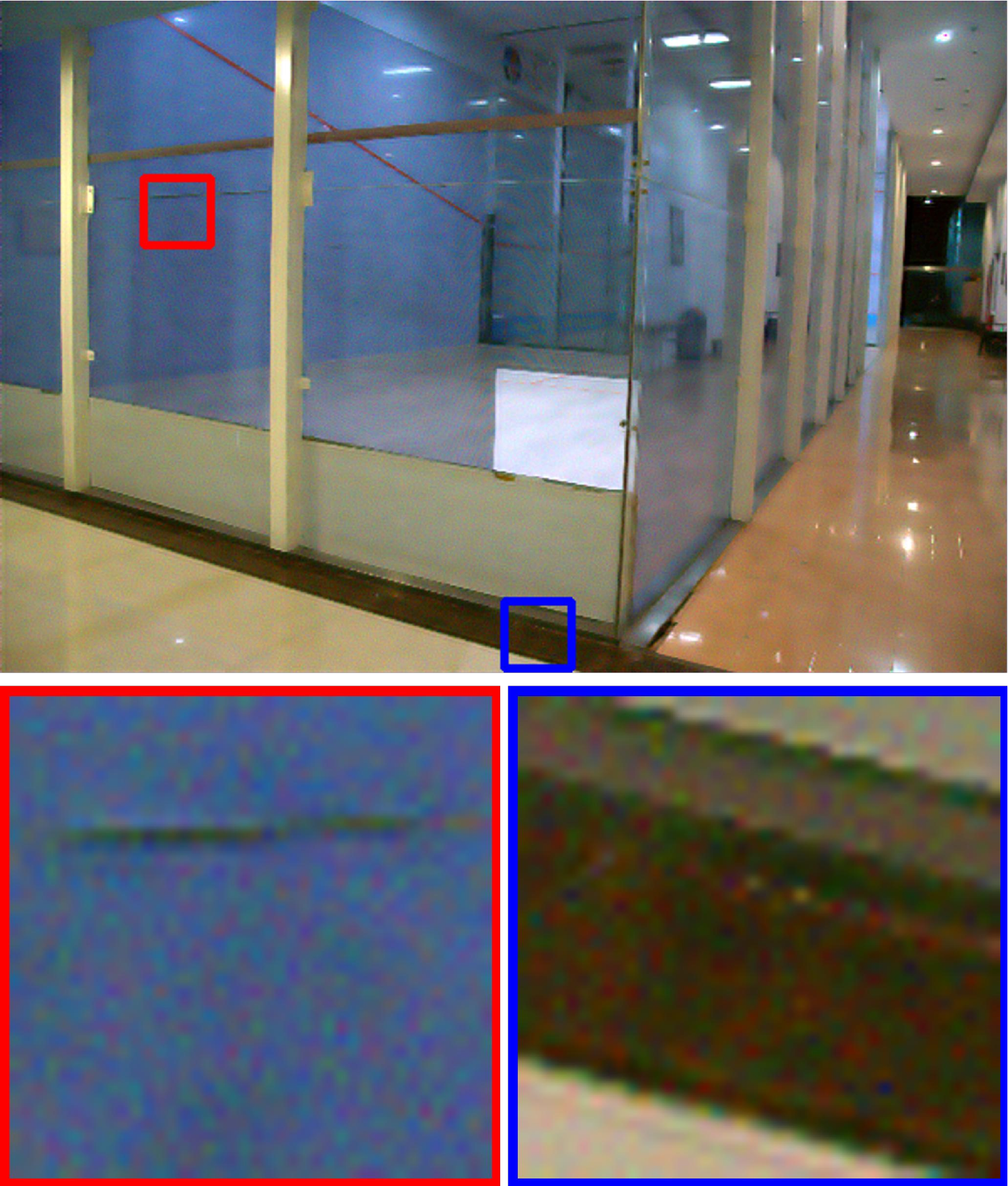}}
    \centerline{\small 19.229/0.511}
    \centerline{\small RUAS \cite{RUAS}}
    \vspace{1pt}
    \centerline{\includegraphics[width=\textwidth]{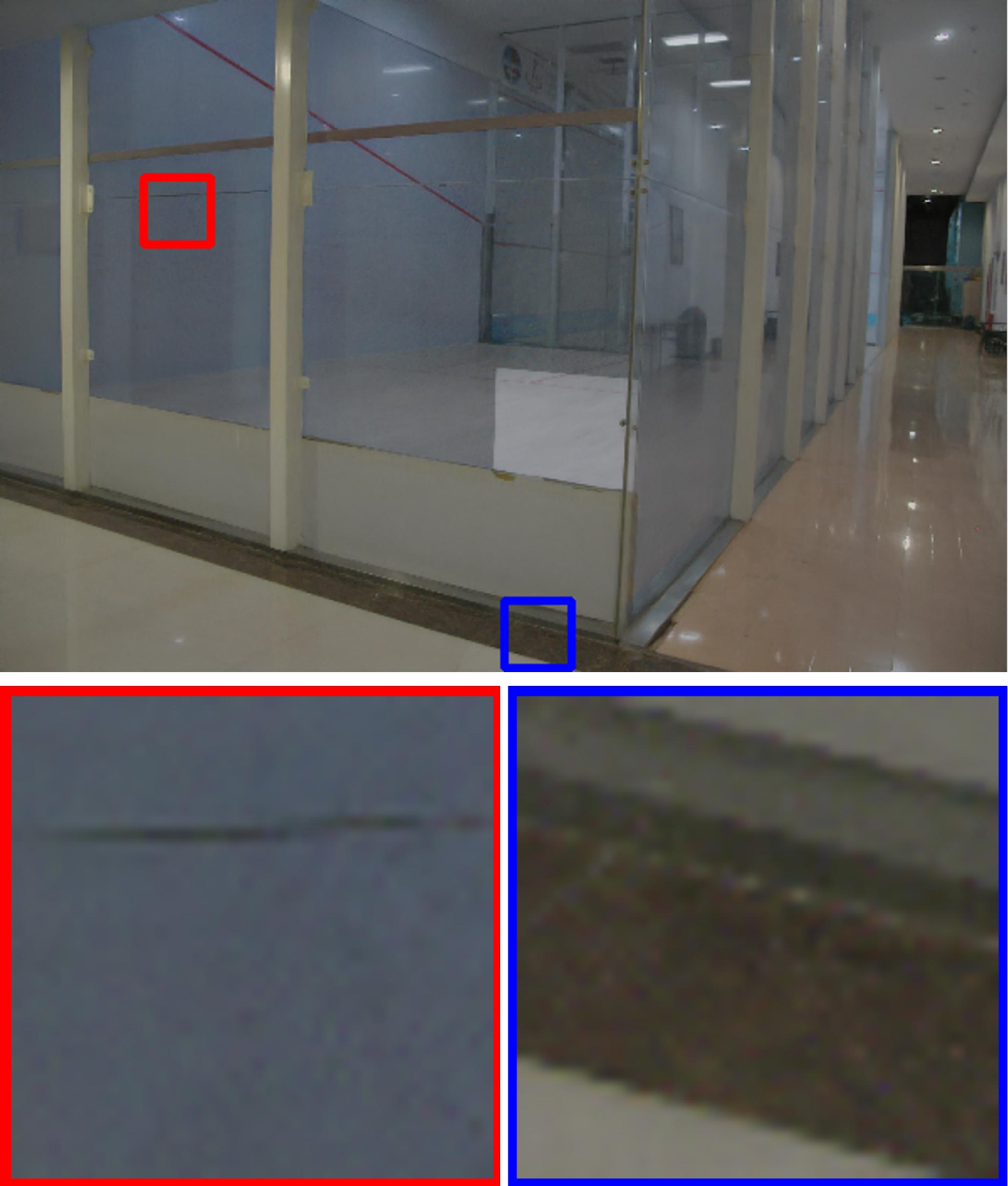}}
    \centerline{\small 31.223/0.895}
    \centerline{\small CIDNet \cite{yan2025hvi}}
\end{minipage}
\hfill
\begin{minipage}[t]{0.19\linewidth}
    \centering
    \vspace{1pt}
    \centerline{\includegraphics[width=\textwidth]{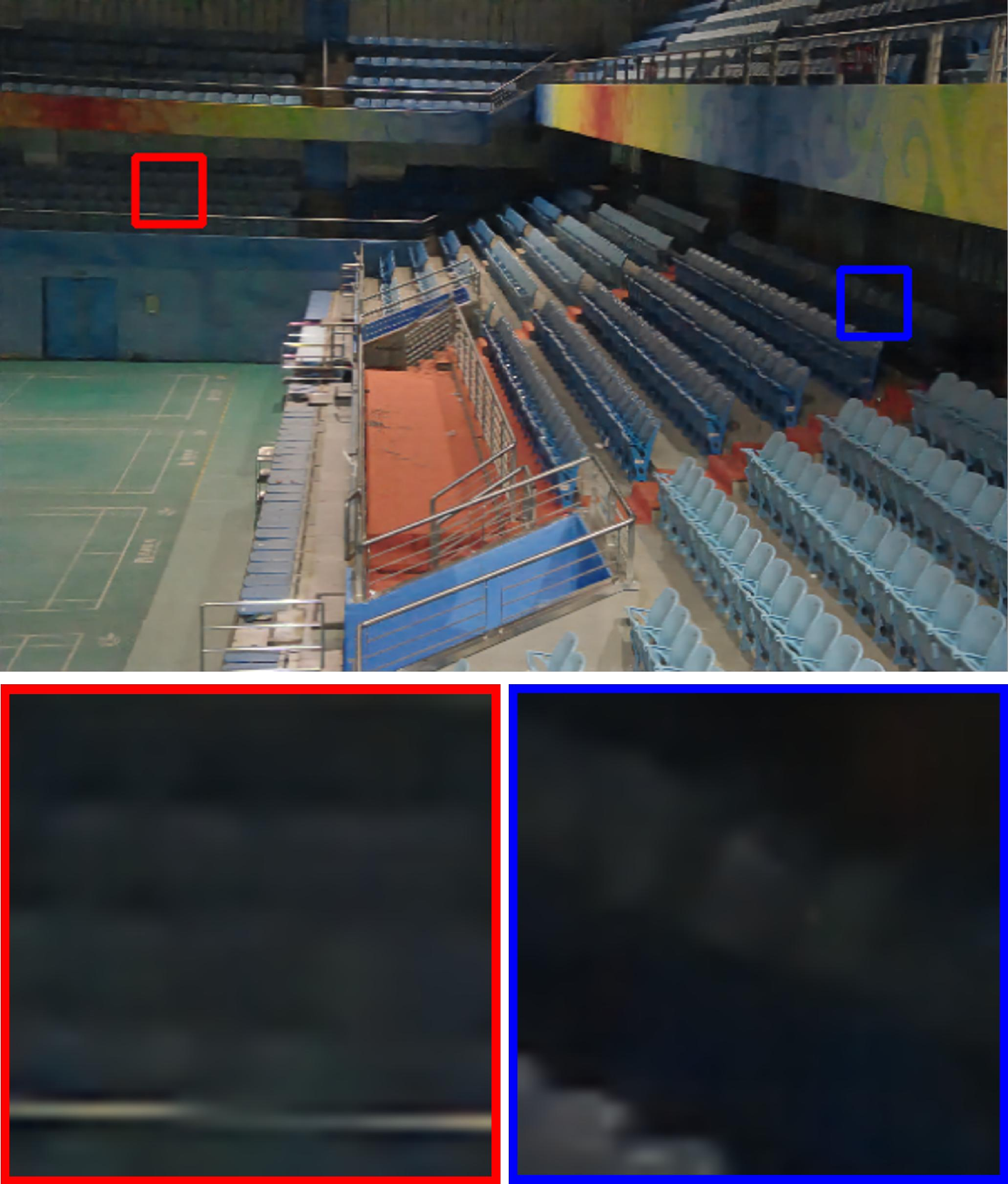}}
    \centerline{\small 23.431/0.768}
    \centerline{\small SNRNet \cite{SNR-Aware}}
    \vspace{1pt}
    \centerline{\includegraphics[width=\textwidth]{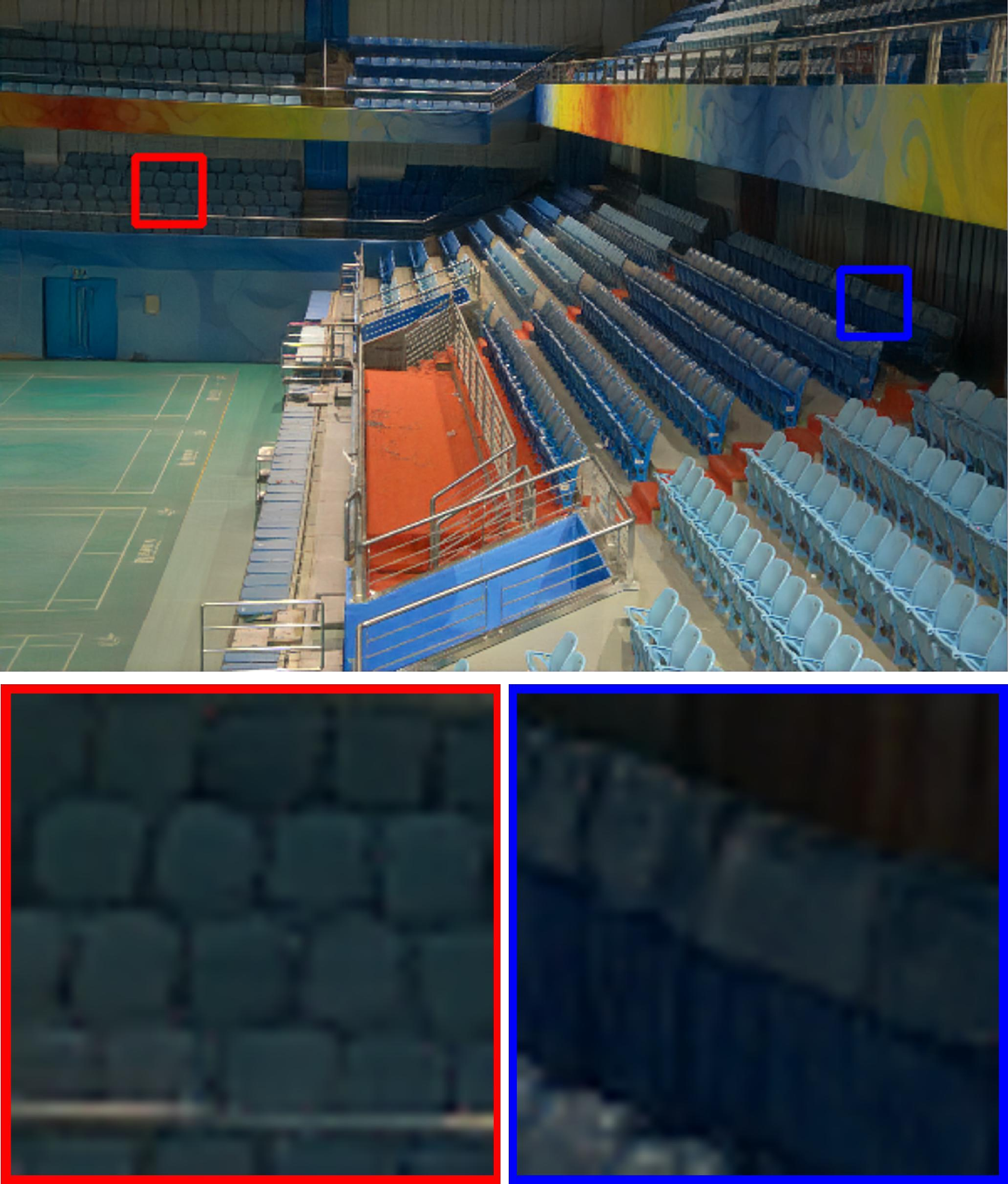}}
    \centerline{\small \textcolor{red}{26.681/0.840}}
    \centerline{\small HVI-CIDNet+(Ours)}
    \vspace{4pt}
    \centerline{\includegraphics[width=\textwidth]{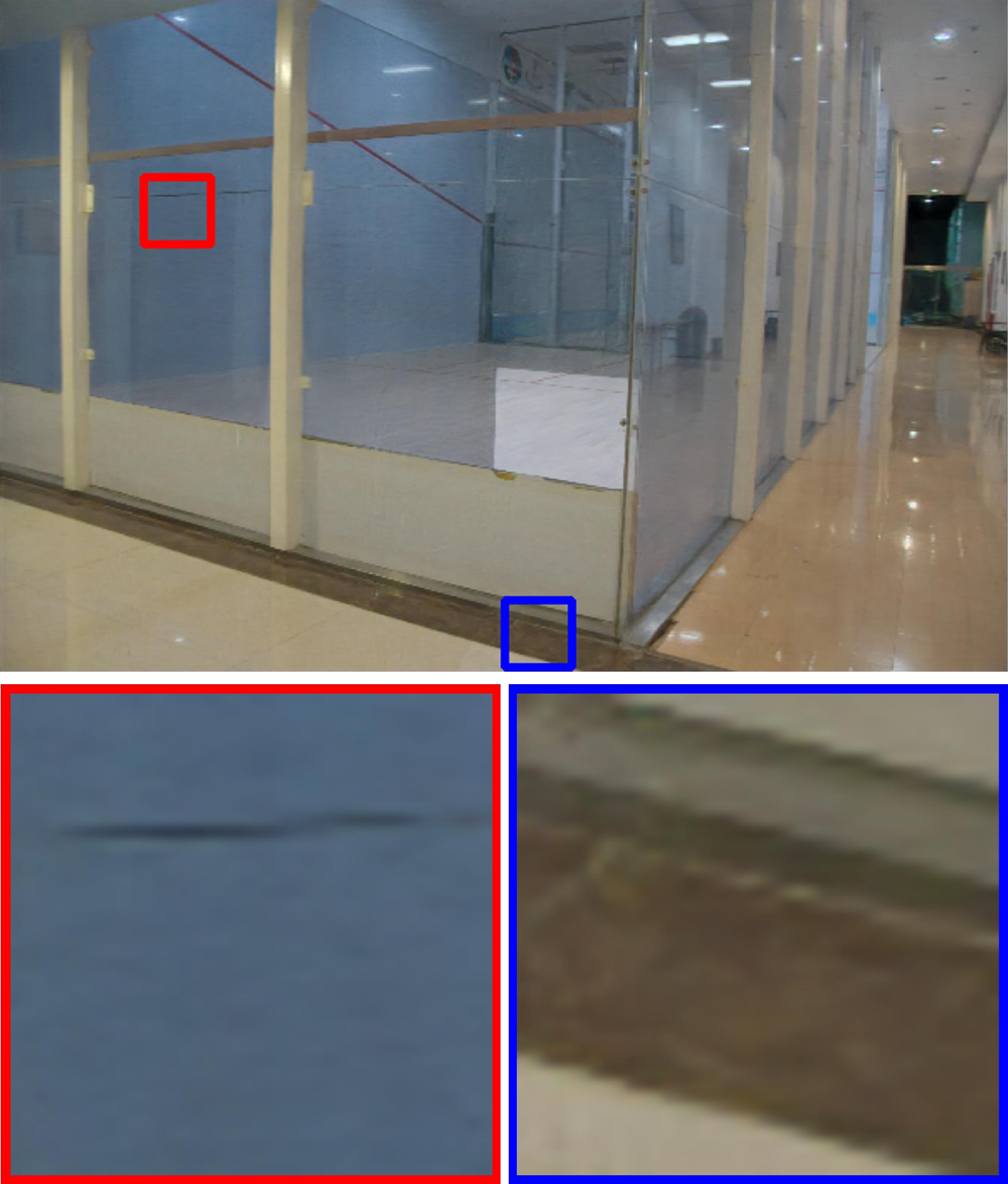}}
    \centerline{\small 21.092/0.882}
    \centerline{\small SNRNet \cite{SNR-Aware}}
    \vspace{1pt}
    \centerline{\includegraphics[width=\textwidth]{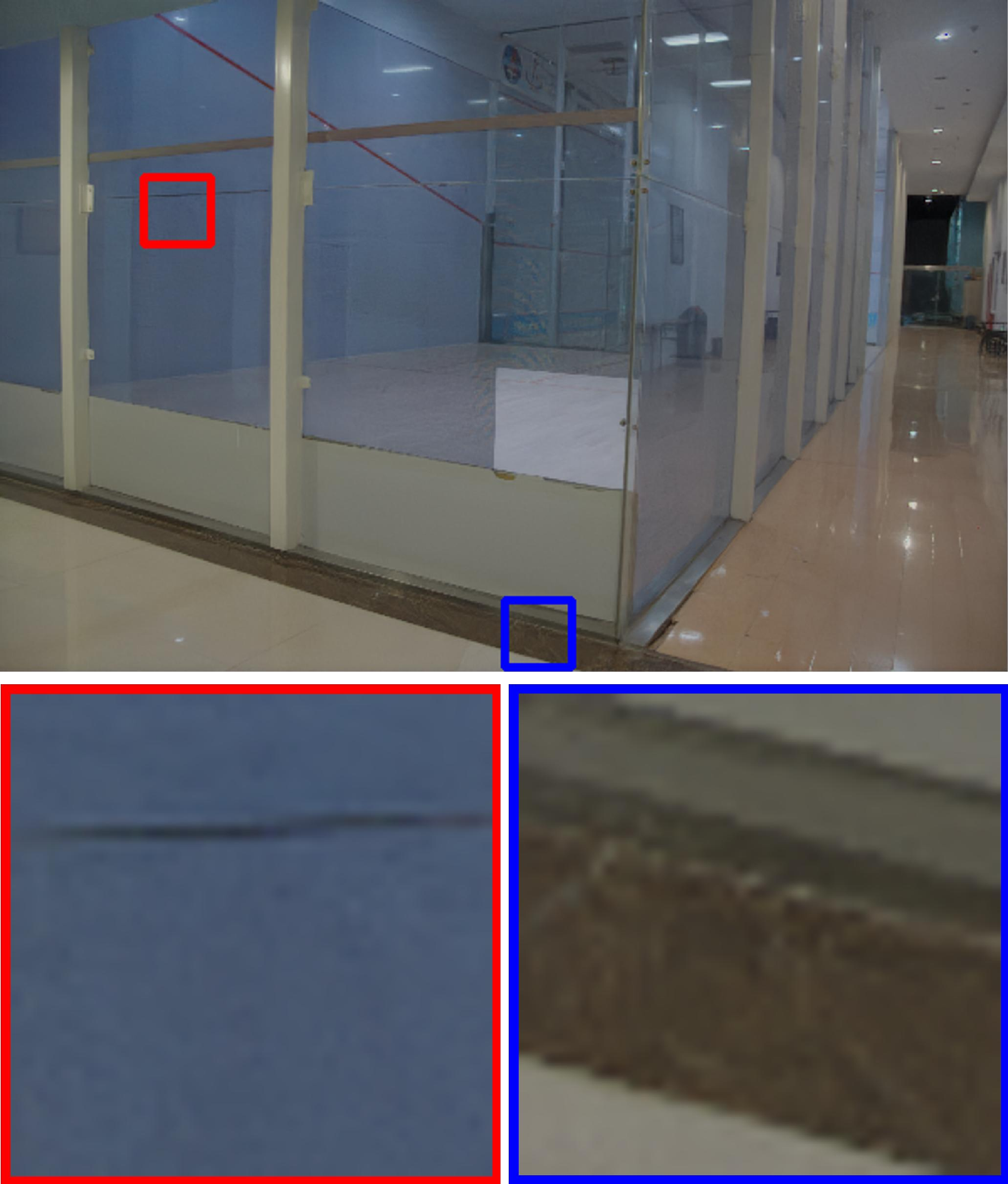}}
    \centerline{\small \textcolor{red}{33.501/0.912}}
    \centerline{\small HVI-CIDNet+(Ours)}
\end{minipage}
\hfill
\begin{minipage}[t]{0.19\linewidth}
    \centering
    \vspace{1pt}
    \centerline{\includegraphics[width=\textwidth]{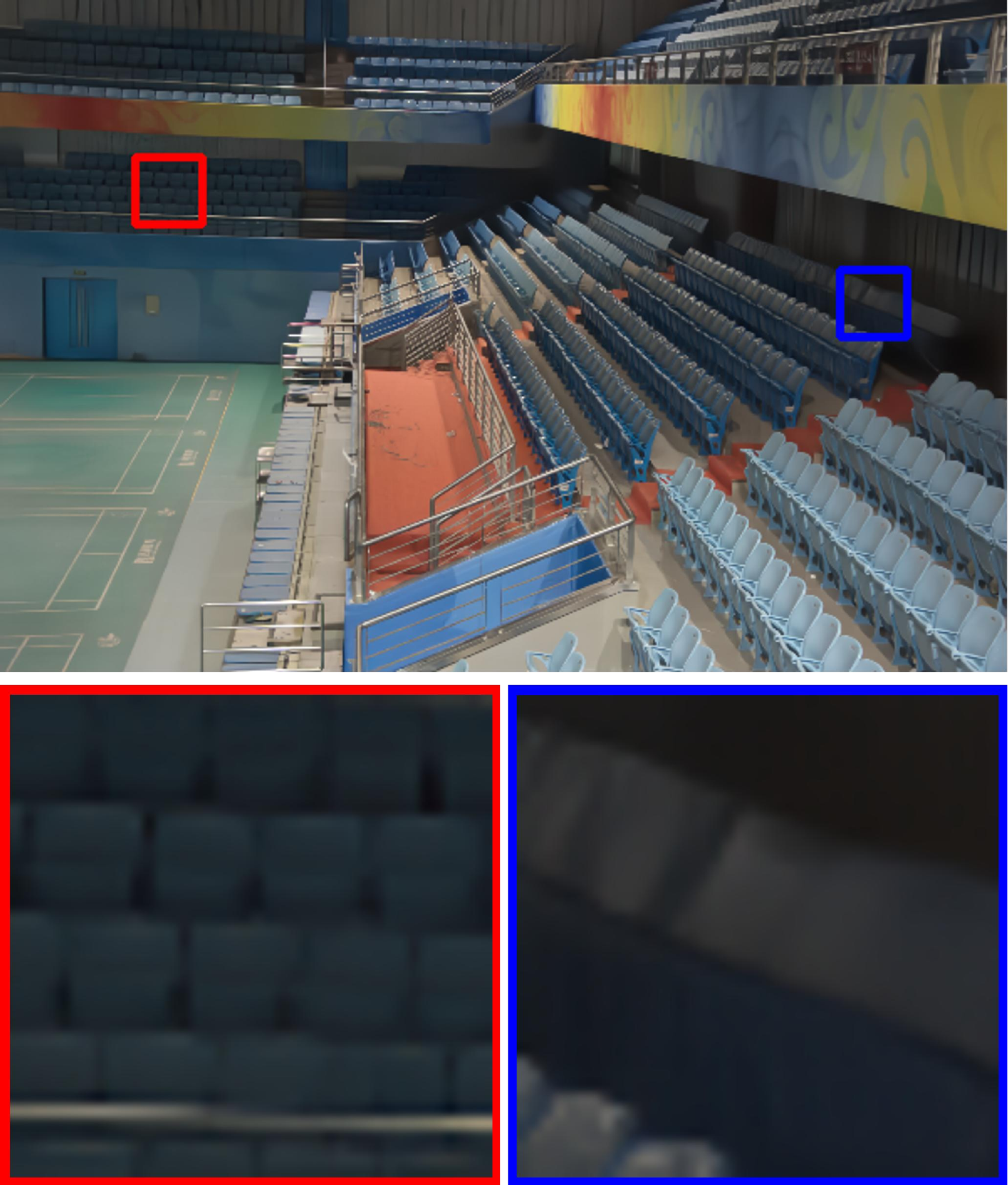}}
    \centerline{\small 24.103/0.828}
    \centerline{\small LLFlow \cite{LLFlow}}
    \vspace{1pt}
    \centerline{\includegraphics[width=\textwidth]{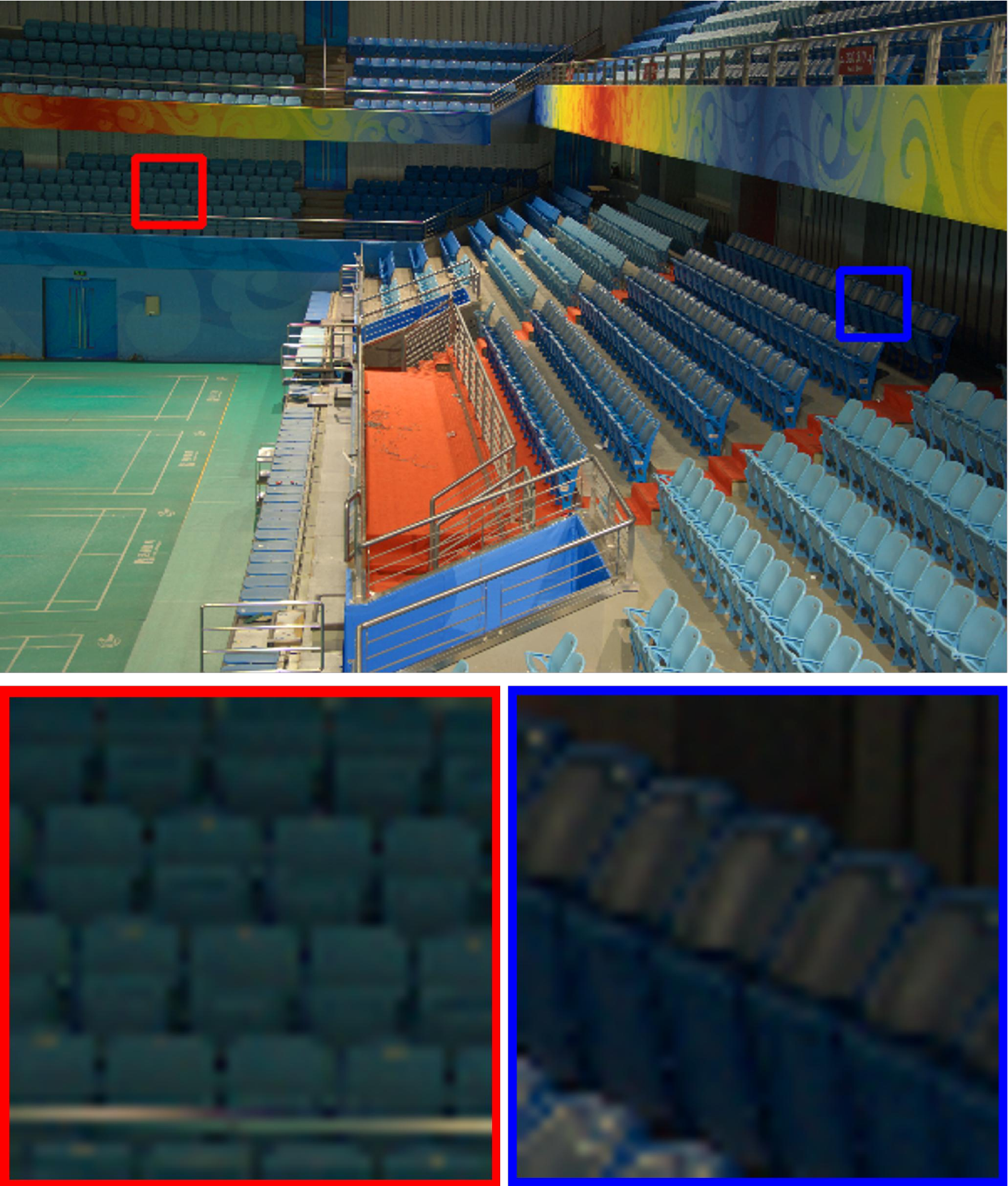}}
    \centerline{\small PSNR$\uparrow$/SSIM$\uparrow$}
    \centerline{\small Ground Truth}
    \vspace{4pt}
    \centerline{\includegraphics[width=\textwidth]{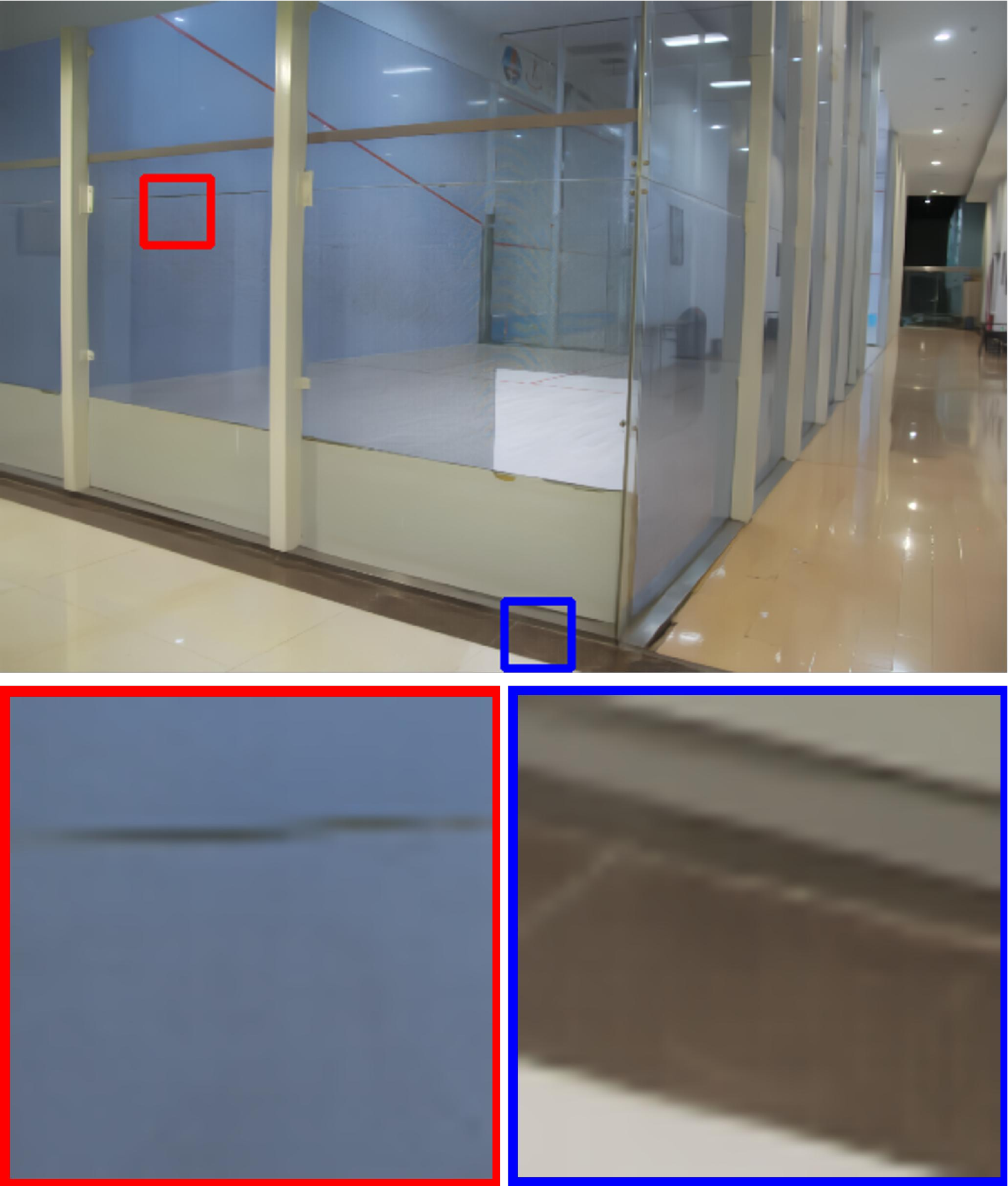}}
    \centerline{\small 16.980/0.864}
    \centerline{\small LLFlow \cite{LLFlow}}
    \vspace{1pt}
    \centerline{\includegraphics[width=\textwidth]{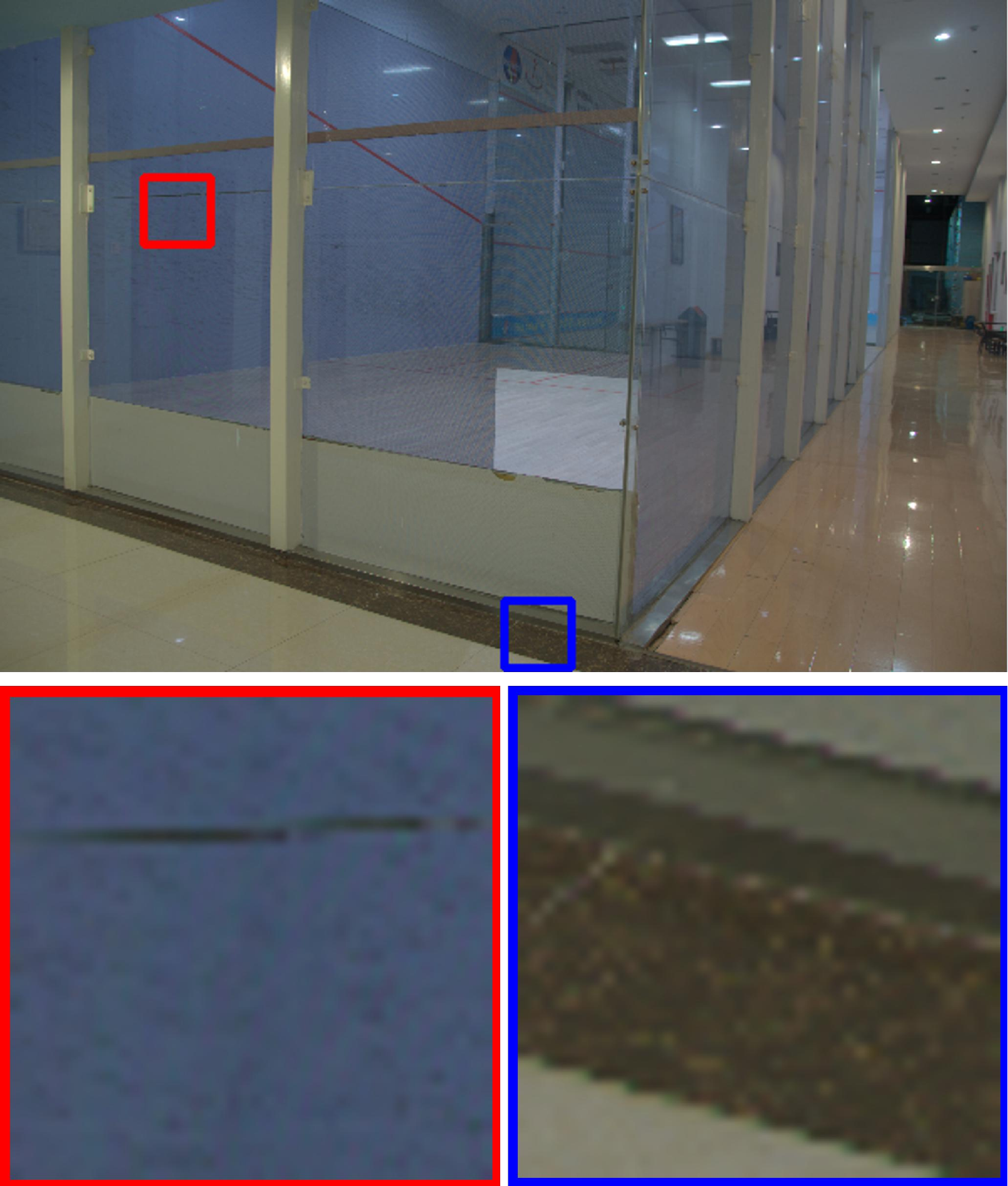}}
    \centerline{\small PSNR$\uparrow$/SSIM$\uparrow$}
    \centerline{\small Ground Truth}
\end{minipage}
\hfill
\vspace{-0.2cm}
 \caption{Visual comparison of the enhanced images yielded by different SOTA methods on LOLv1 (top two rows) and LOLv2 (bottom two rows). Below each image is labeled the PSNR$\uparrow$ and SSIM$\uparrow$ of that measured with Ground Truth, with the best results highlighted in \textcolor{red}{red} color.}
 \label{fig:LOL}
\end{figure*}

\textit{Results on LOL Datasets.}
In Table~\ref{tab:table-LOL}, our method achieves SOTA performance across all metrics on the LOLv1 and LOLv2 datasets. The HVI-CIDNet+ surpasses RetinexFormer (a Retinex-based approach), the leading RGB-based method GSAD (a diffusion-based model), and our previous HVI color space-based work CIDNet.
The subjective results shown in Fig.~\ref{fig:LOL} further illustrate the superiority of HVI-CIDNet+, which delivers higher image quality than RetinexFormer. Although GSAD is capable of capturing detailed textures, it suffers from a significant brightness imbalance. In contrast, HVI-CIDNet+ restores damaged content and mitigate color distortion in extremely dark regions (\textit{\eg} the seating area in the stadium). 
Additionally, it effectively balances enhancement across both information-rich regions and information-rich regions (\textit{\eg} the red region exhibiting colors closer to the Ground Truth image and the blue region showing more accurate details in the interior corridor).
These results demonstrate that HVI-CIDNet+, built upon HVI color space and enhanced by degraded representations, latent semantic priors, and RRB guidance, learns the photometric mapping function more accurately under varying lighting conditions, achieving high-quality image restoration.

\begin{figure*}[!t]
    \centering
    \includegraphics[width=1\linewidth]{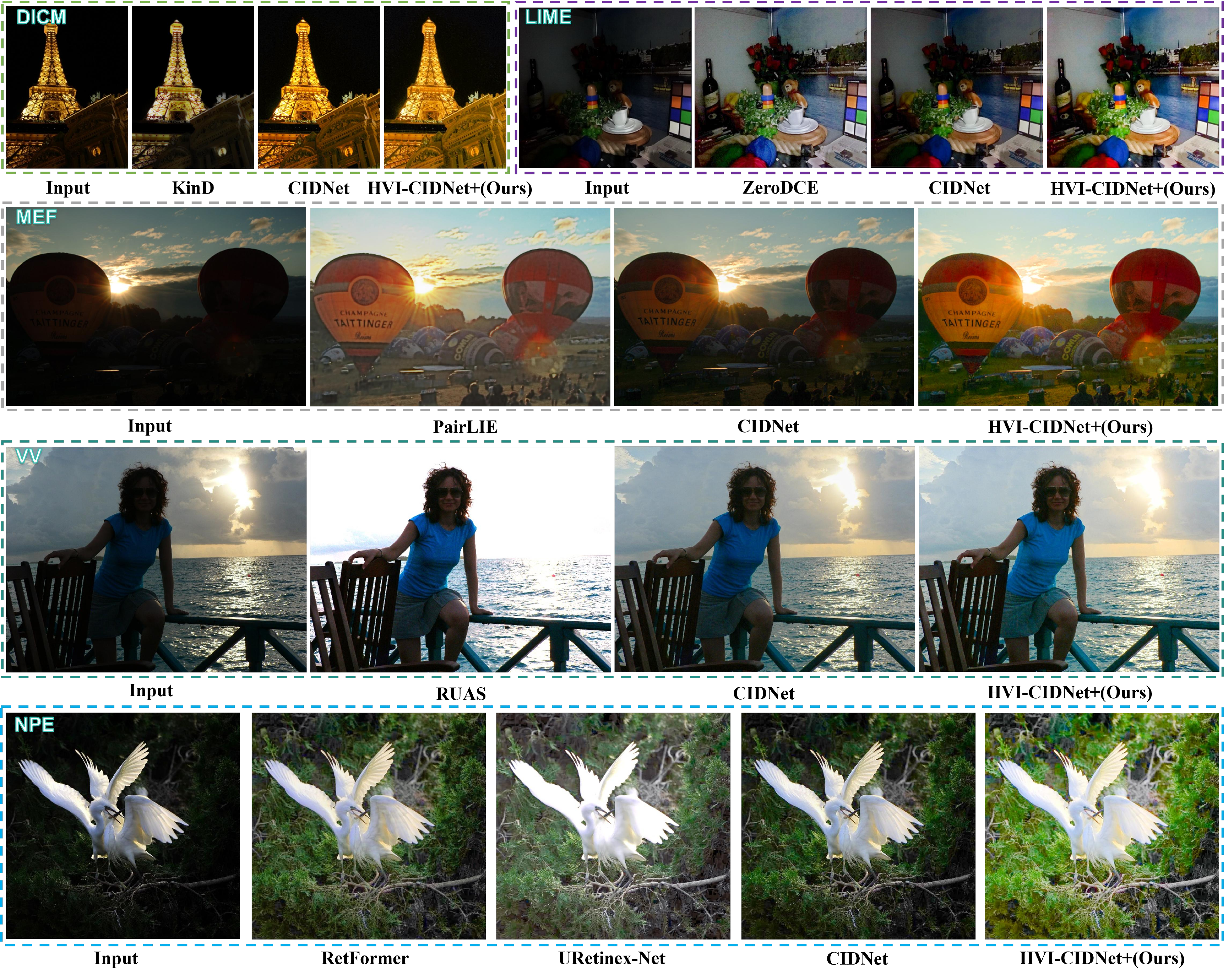}
    \caption{Visual comparison on the five unpaired datasets. Follow RetinexFormer \cite{RetinexFormer}, we select one image in each dataset to compare our method with the other methods. More visual comparison can be found in the supplementary materials.}
    \label{fig:unpaired}
\end{figure*}

\textit{Results on Unpaired Datasets.} 
We evaluate the generalization ability of various methods trained on LOLv1 or LOL-v2 dataset by testing them on unpaired datasets. Since Ground Truth references are unavailable in this setting, we adopt no-reference image quality assessment metrics, specifically BRISQUE and NIQE, to evaluate performance. As shown in Table~\ref{tab:SID}, our method achieves significant improvements over existing approaches in both metrics, indicating better perceptual quality and more natural enhancement results. The visual results shown in Fig.~\ref{fig:unpaired} are more convincing and fair to justify the superiority. These results demonstrate that HVI-CIDNet+ not only performs well in paired settings but also effectively generalizes to unpaired, real-world low-light conditions.

\textit{Results on SICE and Sony-total-Dark.} 
To validate performance under extremely challenging conditions, we 
assess HVI-CIDNet+ on SICE (including Mix and Grad subsets) and Sony-Total-Dark. 
Due to the difficulty of evaluating visual quality in these datasets, where extreme illumination environments can distort human perception, we rely solely on quantitative metrics for assessment. The results, presented in Table~\ref{tab:SID}, show that HVI-CIDNet+ achieves the highest performance on both PSNR and SSIM across the SICE and Sony-Total-Dark datasets. The improvements are particularly notable. 
On Sony-Total-Dark, HVI-CIDNet+ surpasses the LLFlow method by 7.256 dB in PSNR. On SICE, it achieves a gain of 1.713 dB in PSNR over LLFlow. Meanwhile, HVI-CIDNet+ also outperforms our previous work CIDNet \cite{yan2025hvi}.
These gains reflect the extreme darkness of the dataset images, which makes distinguishing details from noise particularly challenging. HVI-CIDNet+ addresses this by employing the intensity collapse function $\mathbf{C}_k$ to maintain an optimal signal-to-noise ratio during training, thus eliminating black noise artifacts.
Furthermore, by leveraging the degraded representations and latent semantic priors, HVI-CIDNet+ better restore damaged content and mitigate color distortion in extremely dark regions,  highlighting the significant potential of the abundant contextual and degraded knowledge from pre-trained VLMs in the LLIE task.

\begin{figure*}[!htp]
\centering
\begin{minipage}[t]{0.02\linewidth}
    \centering
    \vspace{1.23cm}
    \centerline{\small (1)}
    \vspace{2.19cm}
    \centerline{\small (2)}
    
\end{minipage}
\begin{minipage}[t]{0.157\linewidth}
    \centering
        \vspace{1pt}
        \centerline{\includegraphics[width=\textwidth]{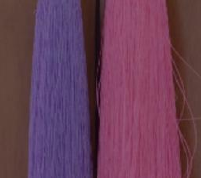}}
        \vspace{5pt}
        \centerline{\includegraphics[width=\textwidth]{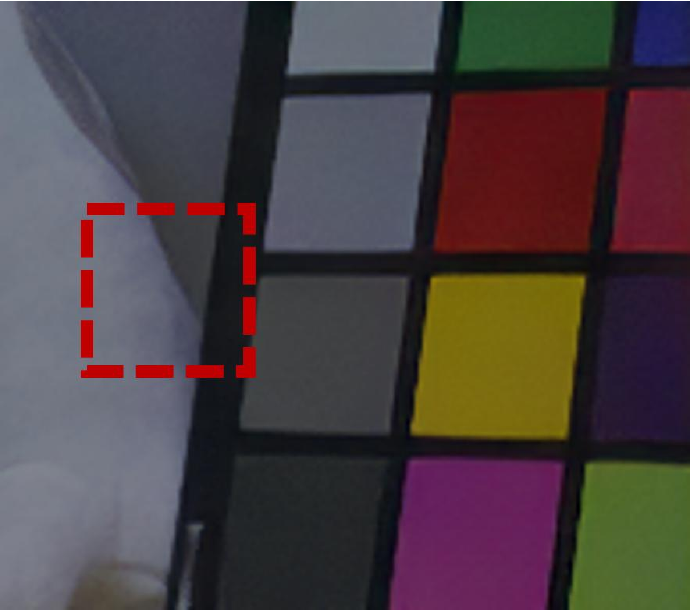}}
    \centerline{\small(a) sRGB}
\end{minipage}
\hfill
\begin{minipage}[t]{0.157\linewidth}
    \centering
        \vspace{1pt}
        \centerline{\includegraphics[width=\textwidth]{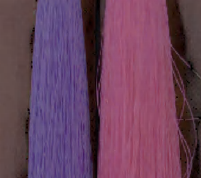}}
        \vspace{5pt}
        \centerline{\includegraphics[width=\textwidth]{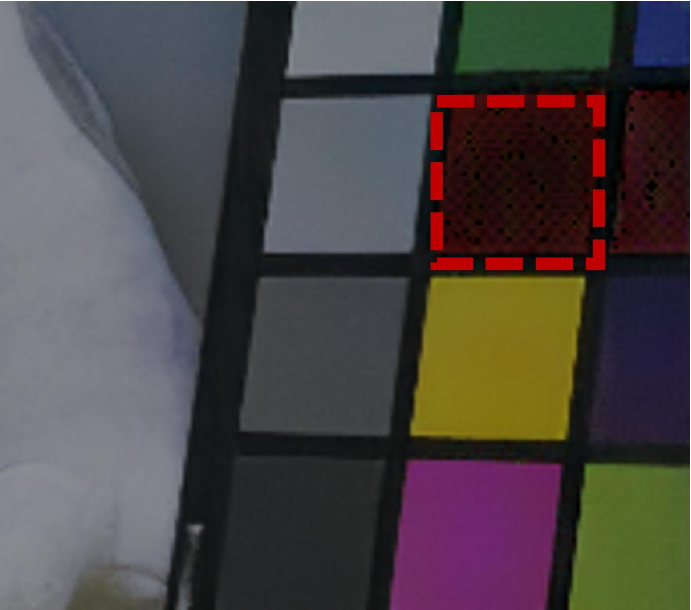}}
    \centerline{\small(b) HSV}
\end{minipage}
\hfill
\begin{minipage}[t]{0.157\linewidth}
    \centering
        \vspace{1pt}
        \centerline{\includegraphics[width=\textwidth]{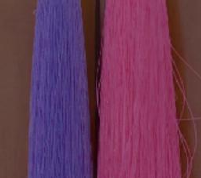}}
        \vspace{5pt}
        \centerline{\includegraphics[width=\textwidth]{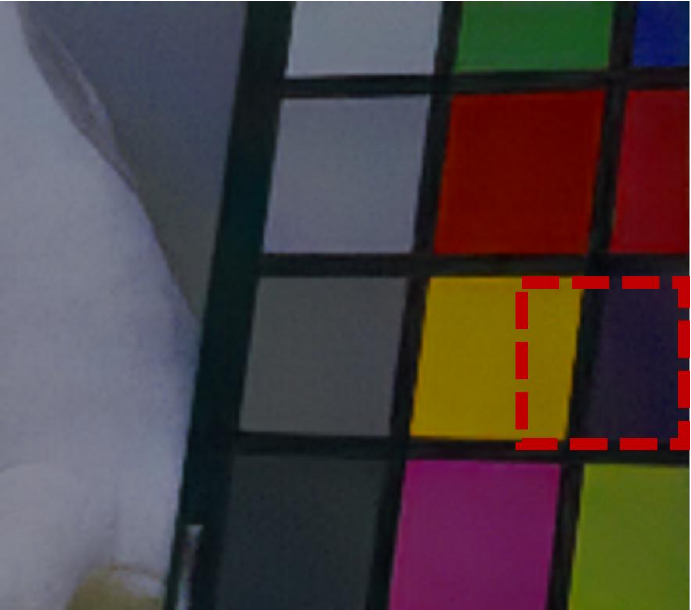}}
    \centerline{\small(c) w/ Polarization}
\end{minipage}
\hfill
\begin{minipage}[t]{0.157\linewidth}
    \centering
        \vspace{1pt}
        \centerline{\includegraphics[width=\textwidth]{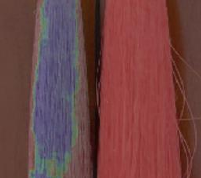}}
        \vspace{5pt}
        \centerline{\includegraphics[width=\textwidth]{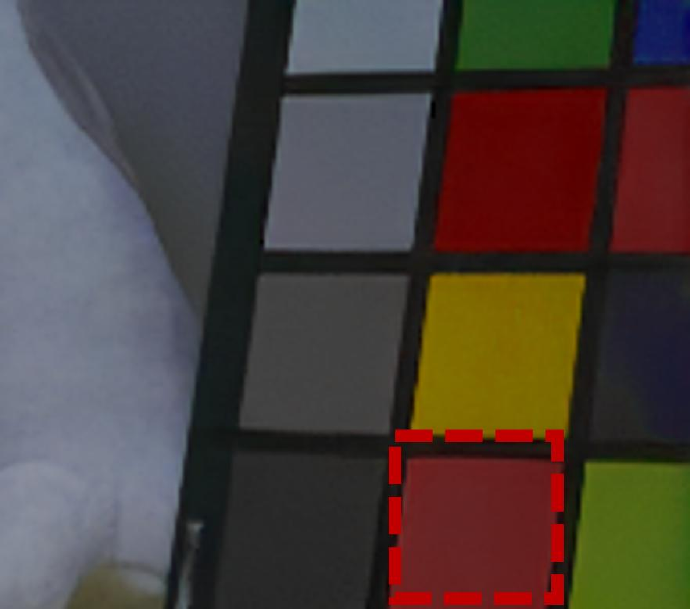}}
    \centerline{\small(d) w/ $\mathbf{C}_k$}
\end{minipage}
\hfill
\begin{minipage}[t]{0.157\linewidth}
    \centering
        \vspace{1pt}
        \centerline{\includegraphics[width=\textwidth]{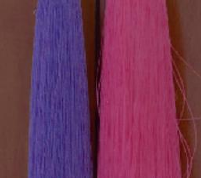}}
        \vspace{5pt}
        \centerline{\includegraphics[width=\textwidth]{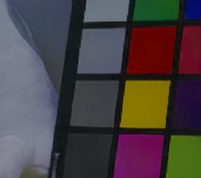}}
    \centerline{\small(e) HVI}
\end{minipage}
\hfill
\begin{minipage}[t]{0.157\linewidth}
    \centering
        \vspace{1pt}
        \centerline{\includegraphics[width=\textwidth]{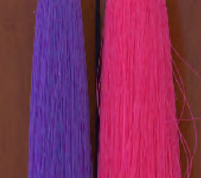}}
        \vspace{5pt}
        \centerline{\includegraphics[width=\textwidth]{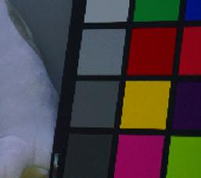}}
    \centerline{\small(f) Ground Truth}
\end{minipage}
\caption{Abaltion results on LOLv1 dataset for five different color spaces used by HVI-CIDNet+. The area outlined by the \textcolor{red}{red} dashed box shows obvious enhancement issues. (Zoom in for the best view.)} 
\label{fig:hvi2}
\end{figure*}
\textit{Generalizing HVI to Other LLIE Models.}
To further validate the effectiveness of the HVI color space, we evaluate its integration with different LLIE models. Specifically, the HVI transform (HVIT) and its inverse mapping (PHVIT) are incorporated as a plug-and-play module into six SOTA methods that operate on sRGB inputs and are not dependent on any specific color space assumptions. The results show consistent improvements in the PSNR, SSIM, and LPIPS metrics when the models are applied in the HVI color space compared to their original sRGB-based counterparts in Table~\ref{tab:HVI}. Notably, the GSAD method exhibits the most substantial gain, with a PSNR improvement of 3.562 dB, which demonstrates not only the generalizability of HVI color space to various sRGB-based methods but also its general effectiveness as a color space for the LLIE task.
\subsection{Ablation Study}
\begin{figure*}[!htp]
\centering
\begin{minipage}[t]{0.136\linewidth}
    \centering
        \vspace{1pt}
        \centerline{\includegraphics[width=\textwidth]{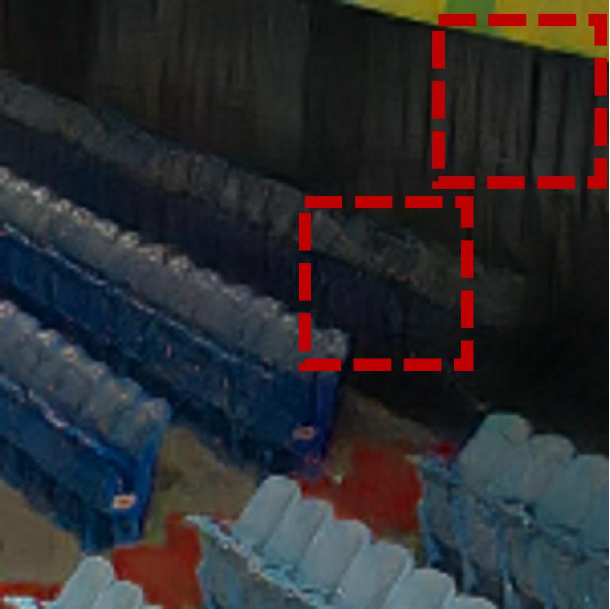}}
    \centerline{\small 27.045/0.701}
    \centerline{\small (a) baseline}
\end{minipage}
\hfill
\begin{minipage}[t]{0.136\linewidth}
    \centering
        \vspace{1pt}
        \centerline{\includegraphics[width=\textwidth]{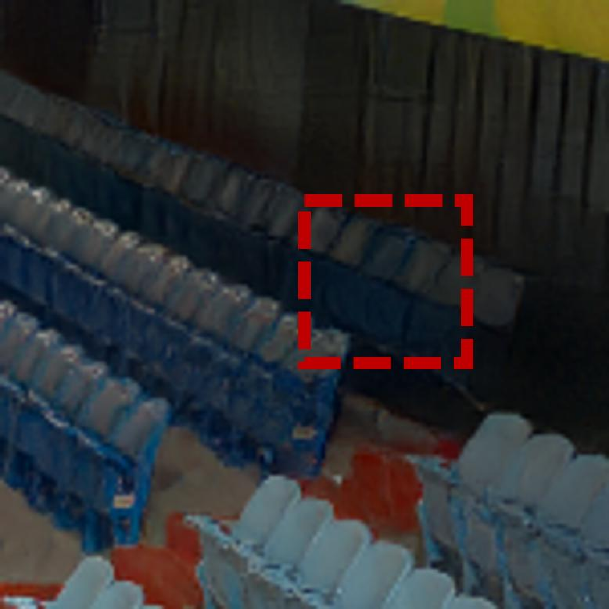}}
    \centerline{\small 27.863/0.740}
    \centerline{\small (b) baseline+$\mathbf{e}_d$}
\end{minipage}
\hfill
\begin{minipage}[t]{0.136\linewidth}
    \centering
        \vspace{1pt}
        \centerline{\includegraphics[width=\textwidth]{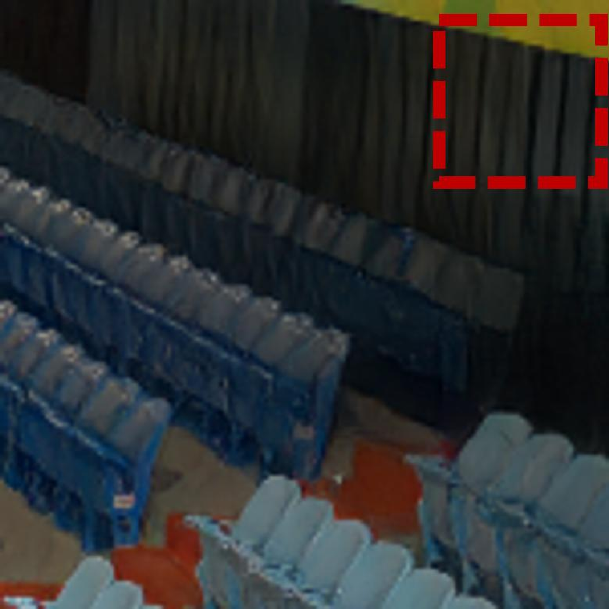}}
    \centerline{\small 27.917/0.739}
    \centerline{\small (c) baseline+$\mathbf{e}_s$}
\end{minipage}
\hfill
\begin{minipage}[t]{0.136\linewidth}
    \centering
        \vspace{1pt}
        \centerline{\includegraphics[width=\textwidth]{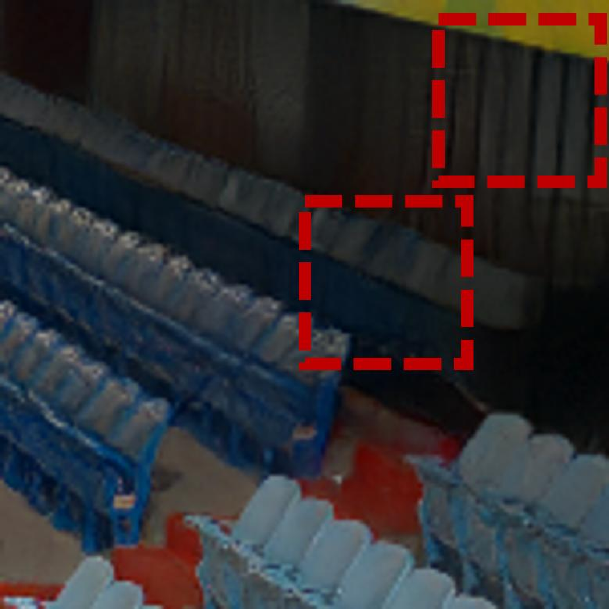}}
    \centerline{\small 28.016/0.744}
    \centerline{\small (d) baseline+$\mathbf{e}_d$+$\mathbf{e}_s$}
\end{minipage}
\hfill
\begin{minipage}[t]{0.136\linewidth}
    \centering
        \vspace{1pt}
        \centerline{\includegraphics[width=\textwidth]{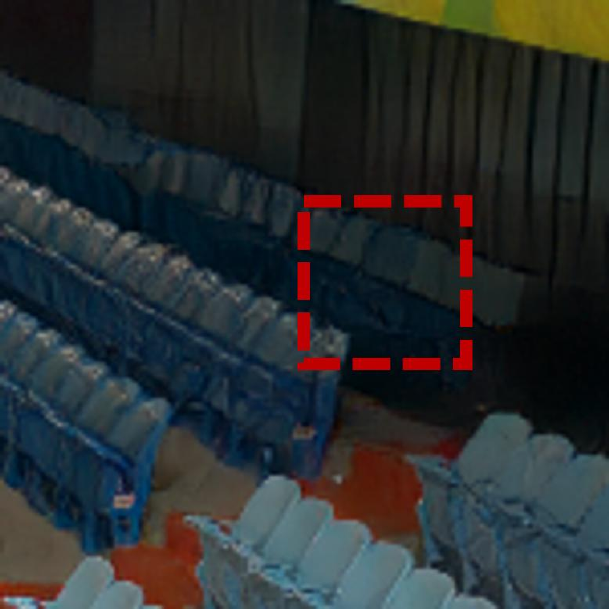}}
    \centerline{\small 27.871/0.734}
    \centerline{\small(e) baseline+RRB}
\end{minipage}
\hfill
\begin{minipage}[t]{0.136\linewidth}
    \centering
        \vspace{1pt}
        \centerline{\includegraphics[width=\textwidth]{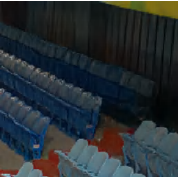}}
    \centerline{\small \textcolor{red}{28.199/0.757}}
    \centerline{\small (f) HVI-CIDNet+}
\end{minipage}
\hfill
\begin{minipage}[t]{0.136\linewidth}
    \centering
        \vspace{1pt}
        \centerline{\includegraphics[width=\textwidth]{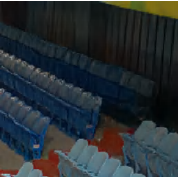}}
    \centerline{\small PSNR$\uparrow$/SSIM$\uparrow$}
    \centerline{\small (g) Ground Truth}
\end{minipage}
\caption{Ablation results of HVI-CIDNet+ by using different structures on LOLv1. Below each image patch is labeled the PSNR$\uparrow$ and SSIM$\uparrow$ of that measured with Ground Truth, with the best results highlighted in \textcolor{red}{red} color. The area outlined by the \textcolor{red}{red} dashed box shows obvious enhancement issues. (Zoom in for the best view.)} 
\label{fig:struct}
\end{figure*}

We validate our HVI color space and network structure in the HVI-CIDNet+ using both quantitative (Table~\ref{tab:ablation}) and qualitative results (Figs. \ref{fig:hvi2} and \ref{fig:struct}). 
For network structure, we perform ablation experiments to assess the impact of degraded representations $\mathbf{e}_d$ and latent semantic priors $\mathbf{e}_s$, as well as the contribution of the Region Refinement Block (RRB).  
All experiments are conducted on the LOLv1 dataset.

\begin{table}[!t]
\centering
\caption{Model ablation. The best PSNR/SSIM$\uparrow$ and LPIPS$\downarrow$ are in \textbf{bolded}.}
\label{tab:ablation}
\vspace{-1.8mm}
\renewcommand{\arraystretch}{1.1}
\resizebox{\linewidth}{!}{

    \begin{tabular}{ll|ccc}
    \hline
         \multicolumn{2}{l|}{\textbf{Metrics}}&	PSNR$\uparrow$&SSIM$\uparrow$&LPIPS$\downarrow$
\\
    \hline
         \multirow{4}{*}{\textbf{Color Space}}&	sRGB	&27.259&0.874& 0.0735
\\
         ~&HSV		&25.857&0.862&0.1007
\\
         ~&HVI (w/ Polarization Only)& 27.788 &0.878 & 0.0795
\\
         ~&HVI (w/ $\mathbf{C}_k$ Only) &27.989  &0.883 &0.0733
\\
\midrule
        \multirow{2}{*}{\textbf{Loss}}&HVI Only&28.102& 0.873&0.0849
\\
        ~&sRGB Only&28.316& 0.892& 0.0620
\\
\midrule
         \multirow{5}{*}{\textbf{Structure}}&CIDNet Baseline~\cite{yan2025hvi}&28.075&0.889&0.0650 	
\\
        ~&baseline+$\mathbf{e}_d$ &28.534&0.891&0.0633
\\
        ~&baseline+$\mathbf{e}_s$ &28.363&0.890&0.0616
\\
        ~&baseline+$\mathbf{e}_d$+$\mathbf{e}_s$ &28.605&0.892&0.0605
\\
        ~&baseline+RRB &28.453&0.890&0.0591
\\

\midrule
        \multicolumn{2}{l|}{\textbf{Full Model (HVI-CIDNet+)}}&\textbf{28.846}&\textbf{0.894}&\textbf{0.0584}
\\
    \hline
    \end{tabular}
    }
\end{table}

\textit{HVI Color Space.} 
After converting images from the sRGB color space to HSV color space, all three evaluation metrics significantly deteriorate as shown in Table~\ref{tab:ablation}. We attribute this deterioration to two factors. On the one hand, the HSV color space introduces red and black noise, which can cause severe artifacts in the enhancement of red- or black-dominated images. On the other hand, images of the LOLv1 dataset contains many red-dominated regions, further amplifying these artifacts.
Applying either polarization or the intensity collapse function $\mathbf{C}_k$ in the HSV color space yields substantial gains, showing that each operation can partially suppress the red and black noise artifacts. Finally, these issues are effectively alleviated when the polarization and the $\mathbf{C}_k$ are applied together, as shown by the consistent improvement of HVI-CIDNet+ in all metrics.

As illustrated in Fig.~\ref{fig:hvi2}, (1) highlights the impact of red noise on restoration quality in red‑related regions, while (2) demonstrates how black noise disrupts overall image appearance.
It can be seen in Fig. \ref{fig:hvi2} that the enhancement in the sRGB color space results in washed-out appearance and low saturation, as demonstrated by the difference between Figs. \ref{fig:hvi2} (a) and (f).
When using the HSV color space instead of the sRGB color space, restored images suffer from obvious red noise, causing edges of red backgrounds to turn black. This can be observed in the enhancement progress from Fig. \ref{fig:hvi2} (1) (b) to Fig. \ref{fig:hvi2} (1) (e). Additionally, the combined effect of red and black noise significantly impairs the visual quality of dark red region, as shown in Fig. \ref{fig:hvi2} (2) (b). 
After applying polarization in the HSV color space, Fig. \ref{fig:hvi2} (1) (c) shows notable improvement. However, the low brightness and unresolved black noise issues lead to inaccurate colors, as illustrated in Fig. \ref{fig:hvi2} (2) (c). 
Similarly, Fig. \ref{fig:hvi2} (d) still fails to address the red noise problem, leading to unsatisfactory  enhancement results. 
Finally, when both the polarization and the $\mathbf{C}_k$ are incorporated to form the HVI color space, Fig. \ref{fig:hvi2} (e) exhibits higher quality, which closely aligns with the Ground Truth.

\textit{Loss Function.} 
This experiment was designed to evaluate the role and effectiveness of the loss function in the HVI color space.  
As shown in Table~\ref{tab:ablation},  compared to using both HVI and sRGB losses, relying solely on the HVI loss lacks pixel-level spatial consistency constraints, leading to a loss of structural detail in the image and thus lower performance across the three metrics, especially in the LPIPS metric. However, using only sRGB loss is focused on pixel-space enhancement, neglecting the low-light probability distribution in the HVI color space, resulting in undesired color imbalance. Thus, utilizing both losses achieves better results.

\textit{Structure.} 
As shown in Table~\ref{tab:ablation}, when incorporating the $\mathbf{e}_d$, the $\mathbf{e}_s$ or the RRB into the baseline model, the performance significantly outperforms the baseline, which demonstrate the key role of these components in improving the model's effectiveness for the LLIE task.

As illustrated in Figs. \ref{fig:struct} (a), (b), (c), and (d), the incorporating of $\mathbf{e}_d$ in Fig. \ref{fig:struct} (b) enables the network to extract color features without distortion, achieving more accurate color correction. In particular, the red dashed box in Fig. \ref{fig:struct} (b) illustrates seat color that is closer to the Ground Truth reference, whereas the clustered, less accurate seat colors in Fig. \ref{fig:struct}(a) show a loss of fidelity.
Meanwhile, the $\mathbf{e}_s$ injects high-level contextual information into the I-branch, such as object shapes, textures, and spatial relationships, which enhances its ability to reconstruct damaged content in extremely dark regions. Consequently, structural details appear sharper, demonstrating the value of semantic guidance for content restoration. As shown in Fig. \ref{fig:struct} (c), the wall exhibits better clear structure and details than the corresponding region in Fig. \ref{fig:struct} (a).
When both the $\mathbf{e}_d$ and the $\mathbf{e}_s$ add to the baseline model, Fig. \ref{fig:struct} (d) demonstrates a significant improvement in visual quality. The PSNR increases by 0.971 dB and SSIM by 0.043. These results emphasize the potential of $\mathbf{e}_d$ and $\mathbf{e}_s$ in achieving content restoration and mitigating color distortion in extremely dark region.
After introducing the RRB, HVI-CIDNet+ can address severe degradation of content and details caused by uneven brightness in low-light images. In the visualization results, as shown in Fig. \ref{fig:struct} (e), the content and detailed texture of the upper seat area is better preserved, no longer losing its sense of hierarchy due to excessive smoothness. RRB achieves an balance between content and details by distinguishing between information-scarce regions and information-rich regions, effectively enhancing the naturalness and optimizing brightness adjustment.
\section{Conclusion}
In this paper, we propose HVI-CIDNet+ for real-world low-light image enhancement, introducing the novel HVI color space to address color noise artifacts arising from HSV color space. 
Built on HVI color space, HVI-CIDNet+ utilizes dual-branch enhancement network with Prior-guided Attention Block (PAB) and Region Refinement Block (RRB) to restore damaged content and mitigate color distortion under the latent semantic priors and the degraded representations guidance in extremely dark regions. 
Experimental results on 10 benchmark datasets demonstrate that HVI-CIDNet+ outperforms SOTA LLIE methods and generalizes well to real-world low-light conditions, showing its robustness in the LLIE task.
\bibliographystyle{IEEEtran}
\bibliography{mybibfile}
\ifCLASSOPTIONcaptionsoff
  \newpage
\fi
\end{document}